%% file: 2021-spikingarm.tex
\documentclass[letterpaper, 10 pt, conference]{ieeeconf}  

\IEEEoverridecommandlockouts                              

\usepackage[utf8]{inputenc}
\usepackage{elasticrow}
\usepackage{multicol}
\usepackage{verbatim}
\usepackage{eurosym}
\usepackage{siunitx}
\usepackage{printlen}
\usepackage[ruled,vlined]{algorithm2e}
\usepackage{algorithmic}
\usepackage{pgfplots}
\usepackage{amsmath}
\usepackage{amssymb}
\usepackage{mathtools}
\usepackage{upgreek}
\usepackage[nice]{nicefrac}
\usepackage[hidelinks]{hyperref}
\pgfplotsset{compat=1.17}
\usepgfplotslibrary{fillbetween}

\usepackage{accents}

\DeclareRobustCommand{\overstar}[1]{\accentset{\vspace{-0.15ex}\smash\ast}{#1}}

\def\imagepadding{0.2cm}
\def\imagepaddingy{0.15cm}

\renewcommand{\vec}[1]{\mathbf{#1}}

\newcommand\defeq{\stackrel{\mathclap{\normalfont\mbox{\scriptsize def}}}{=}}

\newcommand{\pseudoscale}{\ensuremath{\lambda}}

\makeatletter

\renewcommand*\env@matrix[1][\arraystretch]{%
  \edef\arraystretch{#1}%
  \hskip -\arraycolsep
  \let\@ifnextchar\new@ifnextchar
  \array{*\c@MaxMatrixCols c}}
\makeatother

\title{\LARGE \bf
Many-Joint Robot Arm Control with Recurrent Spiking Neural Networks
}

\author{Manuel Traub$^{1}$, Robert Legenstein$^{2}$, and Sebastian Otte$^{1}$
\thanks{$^{1}$Manuel Traub and Sebastian Otte are with the Neuro-Cognitive Modeling group of the Computer Science Department, University of Tübingen, Sand 14, 72076 Tübingen, Germany {\tt\footnotesize\{manuel.traub,sebastian.otte\}@uni-tuebingen.de}}%
\thanks{$^{2}$Robort Legenstein is with the Faculty of Computer Science and Biomedical Engineering, Graz University of Technology, Inffeldgasse 16b, 8010 Graz, Austria
{\tt\footnotesize robert.legenstein@igi.tugraz.at}; Robert Legenstein was supported by the Austrian Science Fund (FWF) within the ERA-NET CHIST-ERA programme (project SMALL, project number I 4670-N)}
}

\begin{document}

\maketitle




\begin{abstract}
In the paper, we show how scalable, low-cost trunk-like robotic arms can be constructed using only basic 3D-printing equipment and simple electronics. The design is based on uniform, stackable joint modules with three degrees of freedom each. Moreover, we present an approach for controlling these robots with recurrent spiking neural networks. At first, a spiking forward model learns motor-pose correlations from movement observations. After training, intentions can be projected back through unrolled spike trains of the forward model essentially routing the intention-driven motor gradients towards the respective joints, which unfolds goal-direction navigation. We demonstrate that spiking neural networks can thus effectively control trunk-like robotic arms with up to 75 articulated degrees of freedom with near millimeter accuracy. 
\end{abstract}

\section{INTRODUCTION}

Elephant trunks are a fascinating invention of biological evolution, particularly from a technical perspective. Due to their flexibility, strength, and effectiveness they have inspired modern robotics and bionics research a lot. Respective robots that try to mimic biological trunks rely, for instance, on pneumatic systems that operate on entire joint-segments, or use several actuators per joint in order to achieve rotations in the relative 2D-plane \cite{bartow2013novel,rolf2013efficient,bayani2015design}.

Calculating the inverse kinematics for robotic actuators is typically a non trivial task, which becomes even more challenging when confronted with hyper-redundant designs and many degrees of freedom as with trunk-like robotic arms \cite{hannan2003kinematics}. 
Researchers also tried to solve this problem with help of machine learning.  Rolf and Stiel, for instance, used a self-organizing map in order to learn a forward model with motor babbling to infer the movements of a pneumatic trunk-like robot \cite{rolf2013efficient,oja2003bibliography}.
In a similar fashion Otte et al. trained a recurrent neural network forward model and used backpropagation through time (BPTT) in combination with a specialized gradient descent variant in order to infer motor inputs of many-joint robotic arms \cite{otte2016inverse,otte2017inherently,otte2018integrative}. It was shown that the inductive bias of recurrent networks provides significant intrinsic advantages for approximating kinematic forward chains with repeated stages of similar computations.
Note that learning kinematic models from data and, in turn, utilizing them for control, is highly relevant for soft robotics, where exact mathematical forward models cannot be formalized easily.

While these and other related works rely on traditional machine learning techniques and second generation neural networks in particular, the recently emerging field of neuromorphic computing increasingly attracts the interest of robotics researchers. It promises highly energy efficient neural computing and even learning directly on mobile platforms \cite{schuman2017survey,roy2019towards,park2020flexible}. These neuromorphic devices typically implement spiking neural networks (SNNs), referred to as the third generation of neural networks. SNNs are biologically more plausible than second generation neural networks as their activation behavior is closer to biological neurons: instead of continuously propagating real-valued activation patterns, spiking neurons communicate in an event-like manner by means of binary spikes that only occur when necessary \cite{maass1999computing}. Despite of their theoretical advantages, SNNs were assumed to be much harder to train and, usually, to perform worse than their second generation counterparts.
For decades their principle potential remained to be shown in practice. Recent developments, however, herald a turnaround here. 
Besides the promising progress with upcoming neuromorphic hardware (cf. e.g. \cite{brainchip}), there are also methodological advancements. Most relevant here is the research of Bellec et al. \cite{bellec2018long,bellec2019biologically}, who presented a novel recurrent SNN variant---the long short-term memory spiking neural network (LSNN)---that can be trained end-to-end via error gradients and that was shown to be competitive with and to partially even outperform state-of-the-art second generation sequence learners, such as LSTMs \cite{hochreiter1997}, on several benchmark problems.

Here we present a method for robustly controlling trunk-like many-joint robot arms with SNNs. Our method combines the RNN-based motor inference technique from \cite{otte2016inverse} with LSNNs \cite{bellec2018long}. In an LSNN-based forward model that learned motor-pose correlations from generated movement observations, intentions are backprojected  along unrolled spike trains essentially routing the motor gradients towards the responsible joints. We show that we can precisely control robot arms with up to 75 articulated degrees of freedom (DoF), purely driven by spikes, with near millimeter accuracy. To the best of our knowledge, this is the first time ever that robotic arms with this complexity have been controlled using SNNs (cf., e.g.,  \cite{bouganis2010,bing2018}). Furthermore, we constructed two different trunk-like robotic arms, which are not only easy to produce using standard desktop 3D-printers, with a total material cost of less than \EUR{500} per robot, they are also controlled with SNNs. Our research opens a novel perspective on the integration of neuromorphic computing and soft robotics.

\section{3D-PRINTED ROBOTIC TRUNKS}

For this paper we developed and explored two similar design alternatives for robotic trunks. Both were designed from scratch using the open source software OpenSCAD \cite{openscad2009} with the same modular principle in mind, namely, an uniform layout for stackable joints of which each can tilt flexibly. This is in contrast to traditional robotic arm designs, where a single joint has typically one tilt axis and, consequently, provides only one articulated degree of freedom. Our joints, on the other hand, provide three degrees of freedom, that is, two tilt axes and a translation along the vertical axis. 

\begin{figure}[t!]
    \begin{elasticrow}[\imagepadding]
      \elasticfigure{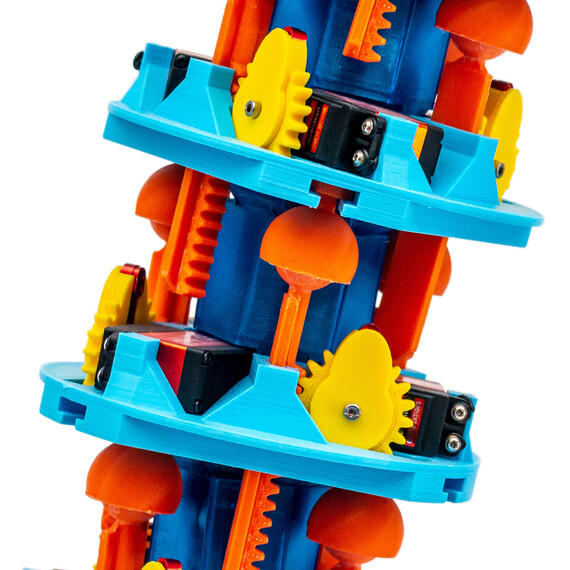}
      \elasticfigure{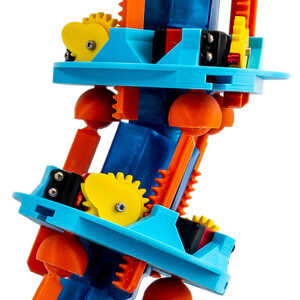}
    \end{elasticrow}
    \vskip\imagepaddingy
    \begin{elasticrow}[\imagepadding]
    \elasticfigure{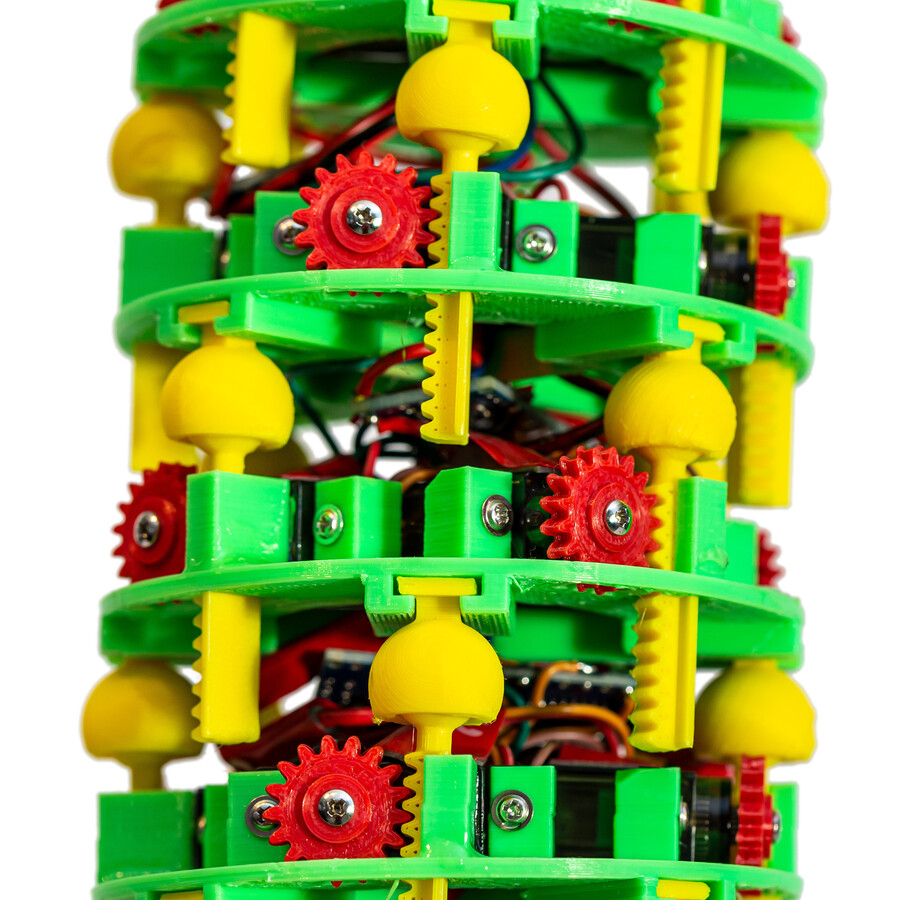}
    \elasticfigure{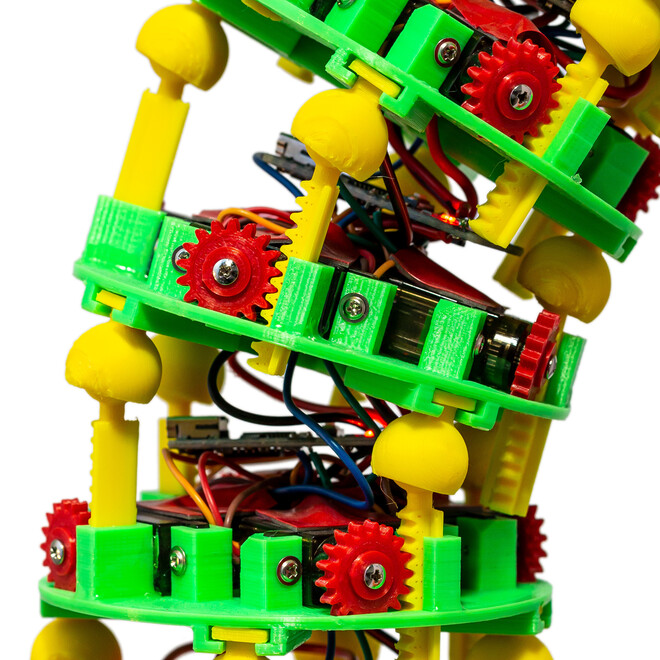}
    \end{elasticrow}
    \caption{Close-up images of the 3-geared joint design (\textbf{top}) and the 4-geared joint design (\textbf{bottom}).
    The movement of an arm joint is controlled with three or four servo motors, respectively, which drive linear gears. These linear gears are in turn connected to the base of the next arm joint via ball-joints and a linear sliding mechanism. In retracted state all linear gears are equally aligned (\textbf{left}). In tilted state the linear gears are shifted reversely to each other (\textbf{right}). The inner electronic parts are either protected by a soft shell (3-geared design) or packed tightly with a rubber band (4-geared design).}
    \label{fig:joint_details}
\end{figure}

The movement is generated via vertically arranged linear gears, driven by servo motors. The two designs have three gears and four gears per joint and are hence referred to as \emph{3-geared design} and \emph{4-geared design}, respectively. \autoref{fig:joint_details} depicts the particular differences in detail. The linear gears are connected through closed ball-joints together with a slider mechanism to the base of the next joint. Note that the ball-joints were printed already in their encapsulated form.

By adjusting the linear gears in the right way, a joint can approach any combination of tilt angles along the $x$ and $y$ axes in the range of $[-16\si{\degree},16\si{\degree}]$ for the 4-geared robot and $[-40\si{\degree},40\si{\degree}]$ for the 3-geared version. Additionally, a joint can alter the distance to the next joint, although the approached angles limit the movable distance and vice versa. For instance, when either the tilt on the $x$ or $y$ axis is at the maximum possible angle, then the distance along the vertical $z$ axis cannot be altered anymore. In case of no tilt, however, the possible range of vertical stretch is $22$\,mm for the 4-geared robot and $55$\,mm for the 3-geared version.

\begin{figure*}[t!]
    \includegraphics[width=\textwidth]{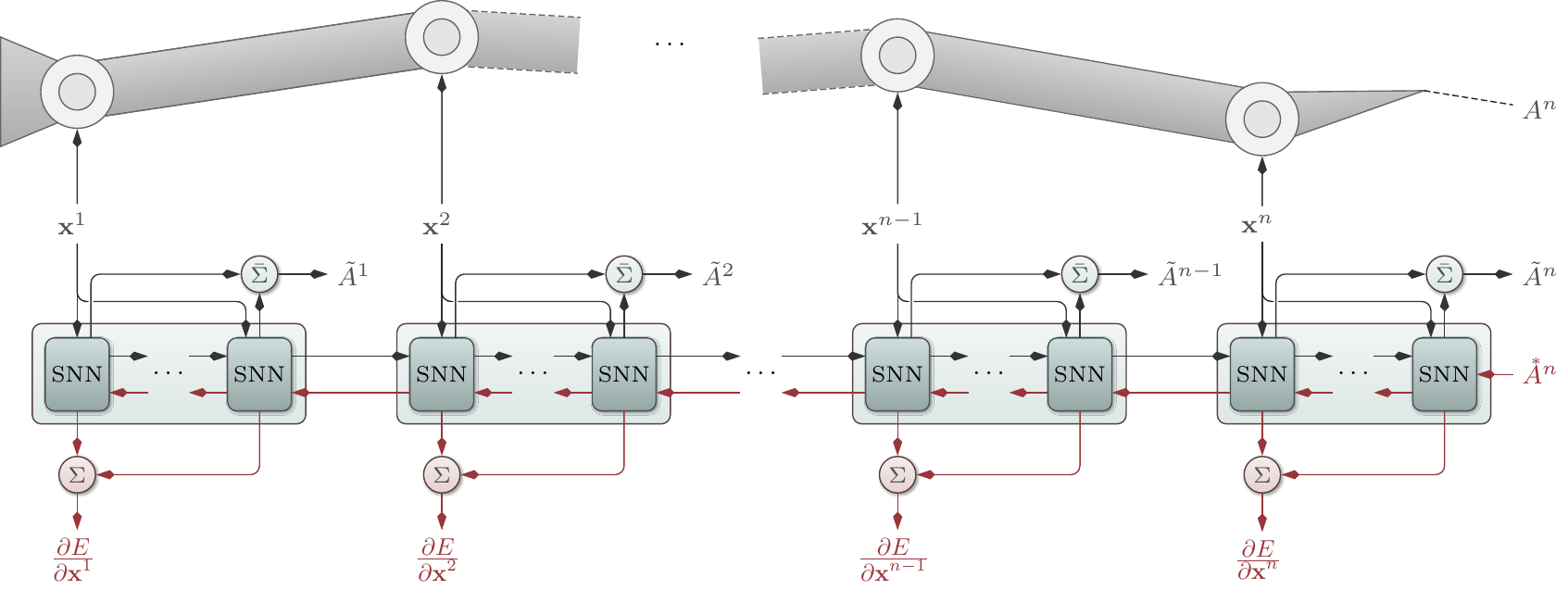}
    \vspace{-0.5cm}
    \caption{Schematic illustration of the SNN forward model and the inference process. During the forward pass (\textbf{black arrows}) the network gets the current joint states $\vec{x}^{k}$ as an input sequence. Additionally, a joint-wise clocking input is given (not shown in the illustration), represented as a one-hot encoded vector indicating the identity of the current joint. For each joint the network performs multiple sub time steps. The associated pose predictions (network outputs) $\tilde{A}^{k}$ are aggregated (averaged) within these time frames.
    For gradient-based motor inference (\textbf{red arrows}) an error is computed based on the current end-effector pose prediction $\tilde{A}^{n}$ and its desired target pose $\overstar{A}^{n}$, where $n$ identifies the end-effetor. This error is then backprojected through the unfolded spike trains of  the forward model and used to derive the input gradient $\frac{\partial E}{\partial \vec{x}^{k}}$, which is aggregated (summed) over the respective time frames.}
    \label{fig:model}
\end{figure*}

The connection of the linear gears with the base of the next joint is accomplished through a combination of ball-joints and linear outward pointing sliders. These sliders are necessary to account for the variation in the difference between the attachment points when approaching different $x$ and $y$ tilt angle combinations. By allowing the linear gear attachments to slide inwards or outwards at the bottom of the next joint, it is possible to approach all different joint states in the 3-geared robot without introducing strain on the linear gears or the connections. In the 4-geared robot there exists a small subset of possible joint states that introduce shear forces on the linear gears, but this is counteracted by the flexibility of the material and in our experiments did not introduce too much stress on the material.

Each joint is controlled with an arduino nano micro-controller, to which all the necessary wires are soldered. There are two different power wirings: one for powering the arduinos with a 12v line, the other one for powering the servo motors with a 7.2v line (3-geared version) and 6v line (4-geared version), respectively. The arduinos are all connected over an inter-integrated circuit (I2C). An additional command arduino, which is also connected to the I2C bus, can be connected to a computer over USB in order to receive and delegate motor commands.

In the 4-geared robot, the wirings are tightly packed using rubber bands or zip ties, while in the 3-geared version an additional capsule printed from flexible material (that can be clicked in place) protects the electronics. 

Most parts of the robots were printed in polylactide (PLA) using a simple fused deposition modeling (fdm) 3D-printer with a bowden extruder. Only the soft capsule and the end-effector of the 3-geared version was printed in thermoplastic polyurethane (TPU) on another fdm 3D-printer with a direct drive extruder.

\section{Methods}

In this section, we first show how forward models for trunk-like robotic arms can be learned from movement observations with recurrent SNNs. Second, we demonstrate our approach for utilizing these SNNs to unfold goal-directed behavior and effectively control the robots via prediction-error induced gradients backprojected through unrolled spike trains.

\subsection{LSNN Forward Model}

The forward model of a robotic arm is implemented using an LSNN, which processes the joint states of the robot in a temporal order, beginning at the base joint in the first processing window up to the last joint before the end-effector in the last processing window. Let $n$ in the following denote the numbers of joints. The procedure is shown in \autoref{fig:model}. In the 4-geared version, the gear states are represented by the $x$ and $y$ tilt angles as well as the relative translation along the vertical $z$ axis for each joint. In the 3-geared version, the network directly gets the normalized position of each linear gear as an input. 
For each joint these values are directly injected as real-valued (input-)currents for a total of 12 simulation time steps. After 5 time steps, the network gets an additional clock input (a one-hot vector with length $n$ indicating the currently ``active'' joint)  and outputs its pose estimation during the remaining 7 time steps. 

The pose estimation $\tilde{A}^{k}=(\tilde{\vec{p}}^{k},\tilde{\vec{q}}^{k})$ is a prediction of the true position $\vec{p}^{k}$ and true orientation $\vec{q}^{k}$ of the currently processed joint $k$. Each position is given as a real-valued 3D vector, which is normalized by the mean inter-joint distance. Each rotation is represented as an unit-quaternion. Using quaternions instead of, e.g., the axis-angle notation, provides a discontinuity-free representation that allows direct meaningful linear-interpolation between different rotations, which is particularly advantageous for our purposes. 

For learning the forward model of a particular robotic arm a data set of random poses is acquired by random sampling from the space of possible joint configurations, additionally including some specific edge cases such as heavily twisted arms. To receive the corresponding end-effector poses, a manually designed mathematical forward model (simulation) or an external feedback system (like, for instance, visual tracking) can be used.

The applied LSNN is a spiking neural network architecture, which we use here with only one single recurrent hidden layer. We directly inject real valued inputs without any explicit spike encoding, which worked best in preliminary experiments. The hidden layer is composed of both leaky integrate and fire (LIF) and adaptive leaky integrate and fire (ALIF) neurons in a one-to-one ratio, following the formulations from Bellec et al. \cite{bellec2018long,bellec2019biologically}:

\subsubsection*{LIF activation}
\begin{equation}\label{eq:leaky_integrate_and_fire}
v_j^{t} = \alpha v_j^{t-1} + \sum_i w_{i,j}^{in} x_i^{t} + \sum_{j'} w_{j',j}^{rec} z_{j'}^{t-1} - z_j^{t-1} v_{thr}
\end{equation}
\begin{equation}\label{eq:lif_spike}
z_j^t = \Theta \left( v_j^t - v_{thr}^t \right)
\end{equation}


\subsubsection*{ALIF activation}
\begin{equation}\label{eq:adap_leaky_integrate_and_fire}
v_j^{t} = \alpha v_j^{t-1} + \sum_i w_{i,j}^{in} x_i^{t} + \sum_{j'} w_{j',j}^{rec} z_{j'}^{t-1} - z_j^{t-1} v_{j,thr}^{t-1}
\end{equation}
\begin{equation}\label{eq:adaption}
a_j^{t} = \rho a_j^{t-1} + z_j^{t-1}
\end{equation}
\begin{equation}\label{eq:adaptive_threshold}
v_{j,thr}^t = v_{thr} + \zeta a^t_j
\end{equation}
\begin{equation}\label{eq:alif_spike}
z_j^t = \Theta \left( v_j^t - v_{j,thr}^t \right)
\end{equation}

The voltage $v_j$ of a neuron is modeled as an exponentially decaying  sum over the weighted inputs (with the leakage rate $\alpha$), which gets reset after a spike in $z_j$ occurs by subtracting the value of the spike threshold. For LIF neurons the threshold $v_{thr}$ is constant. For ALIF neurons an adaptive threshold $v_{j,thr}$ is used, which depends on the individual spiking activity of the neurons. More specifically, it is based on the firing frequency as formulated in Equation~\eqref{eq:adaption} and Equation~\eqref{eq:adaptive_threshold}: the more often a neuron has spiked recently, the less sensible it becomes to new inputs for a while. $\rho$ determines how quickly the neuron forgets its previous spiking activity and $\zeta$ determines how the accumulated activity influences the adaptive threshold. Due to this behavior---and its formal structure in particular---ALIF neurons act as data (as well as gradient) highways able to bridge even large temporal gaps, which is essential when confronted with long data sequences \cite{bellec2018long}. For both LIF and ALIF neurons, the spike output is computed using the non-differentiable Heaviside function denoted by $\Theta$.

The output layer of the network consists of leaky readout neurons, which are modeled according to Equation~\eqref{eq:leaky_integrate_and_fire} without the reset term.
The entire network is trained end-to-end using BPTT using mean squared error (MSE) as loss. Note that BPTT here only works due to the pseudo-derivative applied to the spiking neurons' activation function \cite{bellec2018long} (cf. Equation~\eqref{eq:psudo_lif} and \eqref{eq:psudo_alif} in the next section).

\subsection{Motor Inference Principle}

The inference process (cf. \autoref{fig:model}) requires a sufficiently trained forward model. The current motor inputs (gear states) $\vec{x}^{k}$ of the robot arm are sequentially propagated through the network, which eventually generates the corresponding end-effector pose prediction $\tilde{{A}}^{n}=(\tilde{\vec{p}}^{n}, \tilde{\vec{q}}^{n})$. Through this unrolled spike-based computation chain, we project the error (MSE-loss) between the prediction $\tilde{{A}}^{n}$ and the desired end-effector pose $\overstar{{A}}^{n}=(\overstar{\vec{p}}^{n}, \overstar{\vec{q}}^{n})$ back in time, that is, along the arm back to its base, onto all motor inputs. 

In the following, we formally derive this input gradient. More details on the full derivation of BPTT in LSNNs can be found in \autoref{section:appendix:bptt} in the appendix. Let us refer to the state of a particular neuron $j$ at time step $t$ as \smash{$\textbf{s}_j^t$}. For LIF neurons this neuron state is one-dimensional and only contains the voltage \smash{$v_j^t$}, whereas for ALIF neurons it is two-dimensional and contains the voltage \smash{$v_j^t$} as well as the threshold adaption value \smash{$a_j^t$}:
\begin{equation}\label{eq:hidden_stae_lif}
\vec{s}_{j,LIF}^t \defeq v_j^t
\end{equation}
\begin{equation}\label{eq:hidden_stae_alif}
\vec{s}_{j,ALIF}^t \defeq 
\begin{bmatrix}[1.4]
v_j^t &
a_j^t
\end{bmatrix}^{\top}
\end{equation}

The full gradient calculation requires all components within the network's computation chain to be differentiable. As mentioned earlier, however, the Heaviside function does not fulfill this requirement. 
To overcome this problem, Bellec et al. \cite{bellec2019biologically} introduced a pseudo-derivative \smash{$h_j^t$}
in place for the non-existing derivative of the threshold function:
\begin{equation}\label{eq:psudo_lif}
\frac{\partial z_j^t}{\partial \vec{s}_{j,LIF}^t} \defeq h_{j,LIF}^t = \pseudoscale ~ \max \left( 0, \left\vert\frac{v_j^t - v_{thr}}{v_{thr}}\right\vert \right)
\end{equation}
\begin{equation}\label{eq:psudo_alif}
\begin{split}
\frac{\partial z_j^t}{\partial \vec{s}_{j,ALIF}^t} 
&\defeq 
\begin{bmatrix}[1.4]
1 &
- \zeta
\end{bmatrix}^{\top}
h_{j,ALIF}^t \\
&=
\begin{bmatrix}[1.4]
1 &
- \zeta
\end{bmatrix}^{\top}
\pseudoscale ~ \max \left( 0, \left\vert\frac{v_j^t - v_{j,thr}^t}{v_{thr}}\right\vert \right)
\end{split}
\end{equation}
where the dumping factor $\pseudoscale$ scales the steepness the linear segments. With help of this pseudo-derivative we can calculate the partial derivative of the loss with respect to \smash{$ \textbf{s}_j^t $}:
\begin{equation}\label{eq:delta_error}
\pmb{\delta}_j^t \defeq \frac{\partial E}{\partial \textbf{s}_j^t}
\end{equation}
by applying the chain rule, which is done in the following for LIF and ALIF neurons.
\subsubsection*{LIF state gradient}
\begin{equation}
\begin{split}
\pmb{\delta}_j^t &= \frac{\partial E}{\partial z_j^t} \frac{\partial z_j^t}{\partial v_j^t} + 
\frac{\partial E}{\partial v_j^{t+1}} \frac{\partial v_j^{t+1}}{\partial v_j^t} \\ &= \frac{\partial E}{\partial z_j^t} h_j^t + \pmb{\delta}_j^{t+1} \alpha
\end{split}
\end{equation}
\subsubsection*{ALIF state gradient}
\begin{equation}
\begin{split}
\pmb{\delta}_j^t &= \frac{\partial E}{\partial z_j^t} \frac{\partial z_j^t}{\partial \vec{s}_j^t} + 
\frac{\partial E}{\partial \vec{s}_j^{t+1}} \frac{\partial \vec{s}_j^{t+1}}{\partial \vec{s}_j^t} \\
    &= \frac{\partial E}{\partial z_j^t} 
    \begin{bmatrix}[1.4]
        h_j^t \\
        -h_j^t \zeta
    \end{bmatrix} 
+  
    \begin{bmatrix}[1.4]
        \alpha & 0 \\
        0 & \rho
    \end{bmatrix}
    \begin{bmatrix}[1.4]
       \delta_{j,v}^{t+1} \\
       \delta_{j,a}^{t+1}
    \end{bmatrix} \\
    &= \frac{\partial E}{\partial z_j^t} 
    \begin{bmatrix}[1.4]
        h_j^t \\
        -h_j^t \zeta
    \end{bmatrix} + 
    \begin{bmatrix}[1.4]
        \alpha \delta_{j,v}^{t+1} \\
        \rho \delta_{j,a}^{t+1} 
    \end{bmatrix}
\end{split}
\end{equation}

Using these delta terms we can finally derive the input gradient:
\begin{equation}\label{eq:input_error}
\begin{split}
\frac{\partial E}{\partial x_ i^t} &= \sum_j \frac{\partial \textbf{s}_j^t}{\partial x_i^t} \frac{\partial E}{\partial \textbf{s}_j^t} \\
&= 
\sum_j w_{i,j}^{in} \pmb{\delta}_j^t
\end{split}
\end{equation}  

Note that we aggregate (sum) the input gradient over all sub time steps of the respective time windows of the associated joints. When we repeatedly apply this inference scheme and adapt the inputs by means of a suitable gradient descent technique, we minimize the prediction error, which effectively moves the arm towards the target pose. The movement of the arm over time directly reflects this optimization process.

For the input inference we propose an enhanced version of the AMSGrad optimizer \cite{reddi2019convergence} augmented with a sign dampening (SD) mechanism from Otte et al. \cite{otte2017inherently}. The optimizer, which we refer to as SD-AMSGrad, is shown in \autoref{algo:opti}.
\begin{algorithm}[b!]
   \caption{SD-AMSGrad}
   \label{algo:opti}
\begin{algorithmic}
\FOR{$\tau \gets (1,...,T) $}
    \STATE $\vec{g}_{\tau} \gets \nabla_{\boldsymbol{\uptheta}} f(\boldsymbol{\uptheta}_{\tau})$
    
    \vspace{0.1cm}
    \STATE $\vec{m}_{\tau} \gets \beta_1 \vec{m}_{{\tau}-1} + (1 - \beta_1) \vec{g}_{\tau} $
    \STATE $\vec{v}_{\tau} \gets \beta_2 \vec{v}_{{\tau}-1} + (1 - \beta_2) \vec{g}_{\tau}^2 $
    \STATE $\vec{s}_{\tau} \gets \beta_3 \vec{s}_{{\tau}-1} + (1 - \beta_3) \operatorname{sgn}(\vec{g}_{\tau})$
    \STATE $\vec{v}'_{\tau} \gets \max(\vec{v}'_{{\tau}-1}, \vec{v}_{\tau})$
    
    \vspace{0.1cm}
    \STATE $\hat{\vec{m}}_{\tau} \gets {\vec{m}_{\tau}}({1 - \prod_{n=1}^{\tau} \beta_1})^{-1}$
    \STATE $\hat{\vec{v}}_{\tau} \gets {\vec{v}'_{\tau}}({1 - \prod_{n=1}^{\tau} \beta_2})^{-1}$
    \STATE $\hat{\vec{s}}_{\tau} \gets {\vec{s}_{\tau}}({1 - \prod_{n=1}^{\tau} \beta_3})^{-1}$
    
    \vspace{0.1cm}
    \STATE ${\boldsymbol{\uptheta}}_{{\tau}+1} \gets \boldsymbol{\uptheta}_{\tau} - \eta \hat{\vec{s}}_{\tau}^2 \frac{\hat {\vec{m}}_{\tau}}{\sqrt{\hat {\vec{v}}_{\tau}} + \epsilon}$
\ENDFOR

\vspace{0.4cm}
\STATE
\hspace{-0.4cm}\parbox{8cm}{%
\footnotesize
$f$ is a differentiable objective function. All vector multiplications/divisions,  $\operatorname{sgn}$, $\operatorname{max}$, and $\sqrt{~}$  apply component-wise.
}

\end{algorithmic}
\end{algorithm}
$\vec{m}_{\tau}$ and $\vec{v}_{\tau}$ estimate the first and second moments of the gradient as in Adam \cite{kingma2015}, where $\vec{s}_{\tau}$ implements an exponential moving average reflecting the gradient sign fluctuations. The more the signs of particular gradient components fluctuate, the smaller the respective values in $\vec{s}_{\tau}^2$ become. Afterwards $\vec{s}_{\tau}$ undergoes a zero-bias correction like $\vec{m}_{\tau}$ and $\vec{v}_{\tau}$. The components of the resulting update vector are individually scaled in accordance with their sign damping values in $\vec{s}_{\tau}^2$.

Thus, SD-AMSGrad does not only allow to move all gears with more balanced relative speeds through the use of gradient variance scaling, it also slows down the movement once a gradient components change signs, which is particularly helpful in case the arm overshoots or oscillates. Also through the max variance scaling, SD-AMSGrad tends to pay more attention to the gradient magnitude later in the inference process and therefore behaves more like pure gradient decent, which was shown to be superior to Adam in previous action inference experiments \cite{otte2017inherently}.

In order to account for prediction errors, the end-effector targets can be continuously adjusted according to Equations \eqref{eq:cor1} and \eqref{eq:cor2}, respectively. Instead of using the targets directly, corrected position $\overstar{\vec{p}}_{corrected}^n$ and orientation $ \overstar{\vec{q}}_{corrected}^n$ targets are calculated based on the difference between the predicted pose and the actual (simulated or observed) pose, which then increase the inference accuracy significantly (cf. Figure \ref{fig:correction}). 
\begin{equation}\label{eq:cor1}
\overstar{\vec{p}}_{corrected}^{n} = \tilde{\vec{p}}^n + \gamma_{1} (\overstar{\vec{p}}^n - \vec{p}^n)
\end{equation}
\begin{equation}\label{eq:cor2}
\overstar{\vec{q}}_{corrected}^n = \tilde{\vec{q}}^n + \gamma_{2} (\overstar{\vec{q}}^n - \vec{q}^n)
\end{equation}
In a real world scenario, the difference between the pose estimation and the actual state of the robot could, for instance, be inferred by a visual tracking system.

\subsection{Simulation}
\begin{figure}[b!]
    \begin{elasticrow}[\imagepadding]
      \elasticfigure{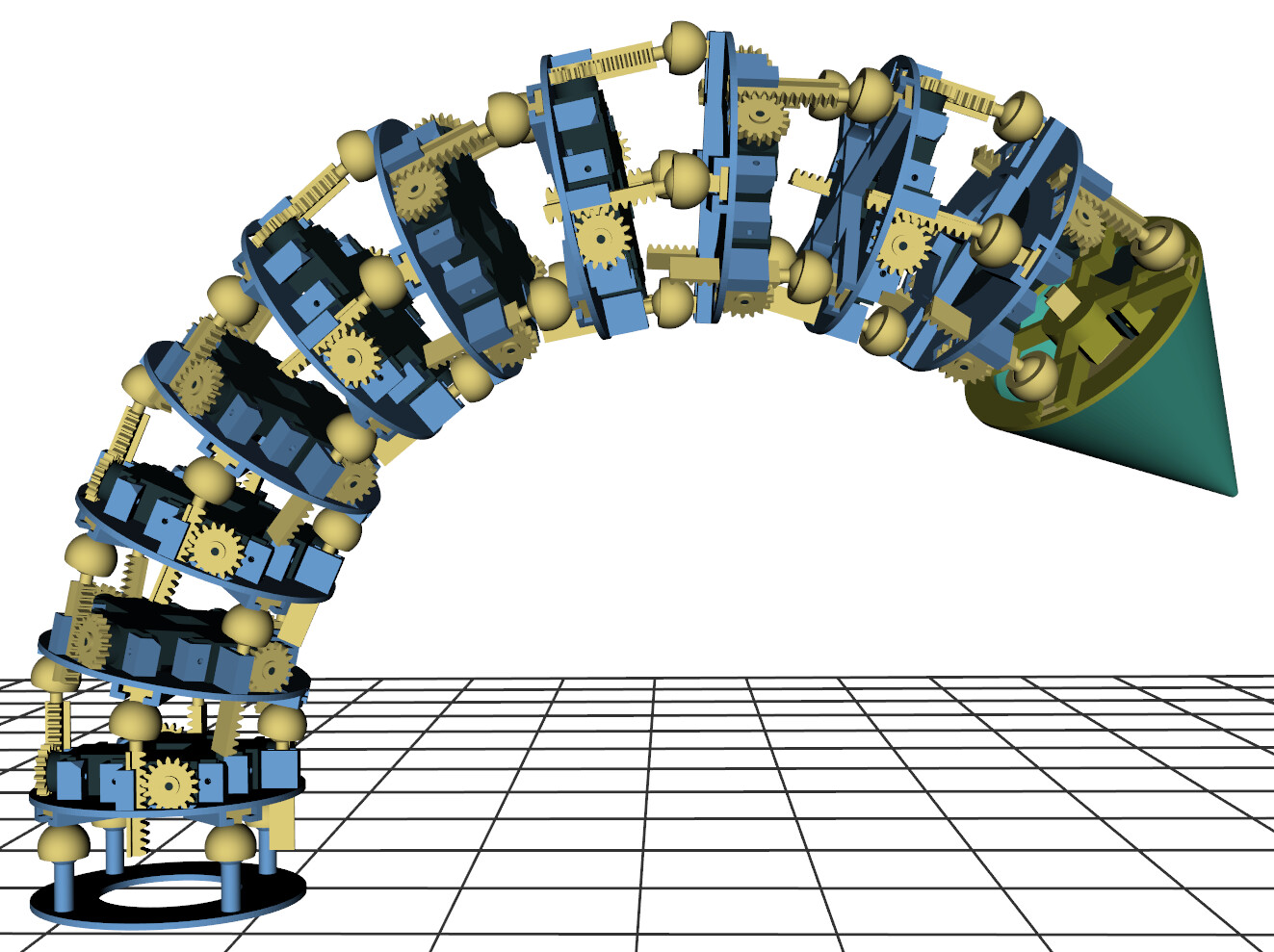}
      \elasticfigure{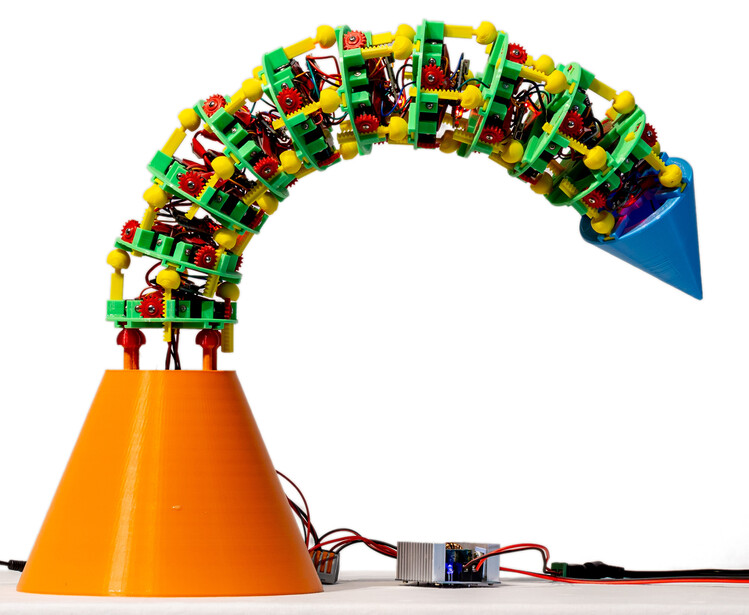}
    \end{elasticrow}
    \caption{Depiction of the 10-joint 4-geared robot arm. The simulation on the left is based on the CAD files from which the real robot on the right was built. The simulation can accurately reproduce all movements possible with the real robot.}
    \label{fig:armv2}
\end{figure}
The simulation of both robotic trunks is based on the CAD parts of the real robots. As an example \autoref{fig:armv2} shows the 4-geared robot arm rendered from the simulation in comparison with its real, 3D-printed counterpart.

We use the simulation to generate the training data, but also as an exact forward model to generate the feedback signal for the target correction. 
While not being physics based, the simulations are on a basic level realistic, meaning that it is ensured that the connections between the joints are realistic and that directly connected CAD parts do not overlap (tolerance is less than $1$ mm).

In order to challenge the networks, two different styles of control strategies are employed. For the 4-geared robot, the networks get the $x$, $y$ tilt angles and the vertical $z$ translation as normalized inputs (range $[-1, 1]$), from which the actual linear gear states are computed. In the 3-geared version, on the other hand, the networks directly receive the state of each linear gear (range $[0, 1]$). 


\section{EXPERIMENTAL SETUP}\label{section:experiments}
The SNN models were trained using Adam with standard parameters \cite{kingma2015}. We used a batch size of 128 and an initial learning rate of 0.001, which was decayed by the factor 0.5 every 10,000 updates. Additionally, a firing rate regularizer as described by Bellec et al. \cite{bellec2019biologically} was applied. The regularizer used a factor of 0.001 in order to allow for a higher firing rate, which seems to be beneficial for this task. The regularizer was decayed in the same way as the learning rate. The training set contains a total of 100,000 samples and the SNNs were trained for a total of 64 epochs, after training the performance was evaluated using a test set of 10,000 samples.

For the evaluation of the prediction and the inference accuracy of the involved models we used two metrics. The first metric as shown in Equation~\ref{eq:dist_error} is the Euclidean distance (reported in millimeters) to quantify the position error.
\begin{equation}\label{eq:dist_error}
    ||E||_2 = \vert\vec{p} - \tilde{\vec{p}}\vert
\end{equation}
The second metric measures the orientation error (reported in degree), as a distance between unit-quaternions representing rotations, as defined in Equation~\ref{eq:rot_error}.
\begin{equation}\label{eq:rot_error}
    E_{rot} = \cos^{-1}( \vert\langle\vec{q},\tilde{\vec{q}} \rangle \vert)
\end{equation}


As an addition to the proposed SD-AMSGrad optimizer, a continuous step size decay mechanism was used, which adjusts the step size $\eta$ (cf. learning rate for training) during the movement of the arm based on the minimum encountered position error as shown in Equation~\eqref{eq:lr}.
\begin{equation}\label{eq:lr}
    \eta^{t+1} = \min\left(\eta^t, \eta^0 \frac{\ln(1 + ||E||_2^t)}{\ln(1 + ||E||_2^0)}\right)
\end{equation}
This mechanism further damps the step size, when the end-effector gets closer to the target. With every new target, we start again with the initial step size.

\section{RESULTS}

\begin{figure*}[t!]
    \begin{elasticrow}[\imagepadding]
        \elasticfigure{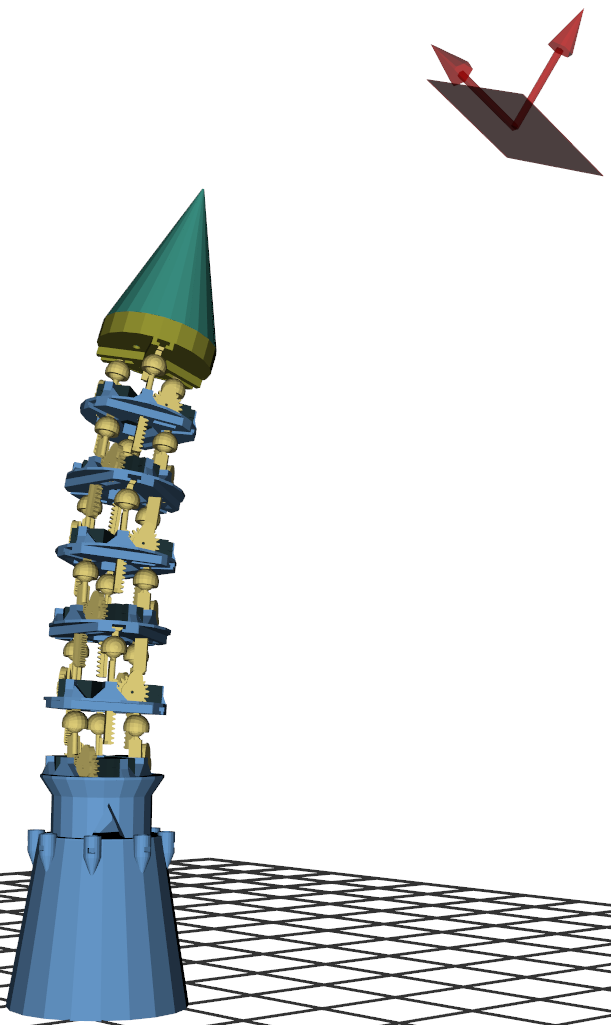}
        \elasticfigure{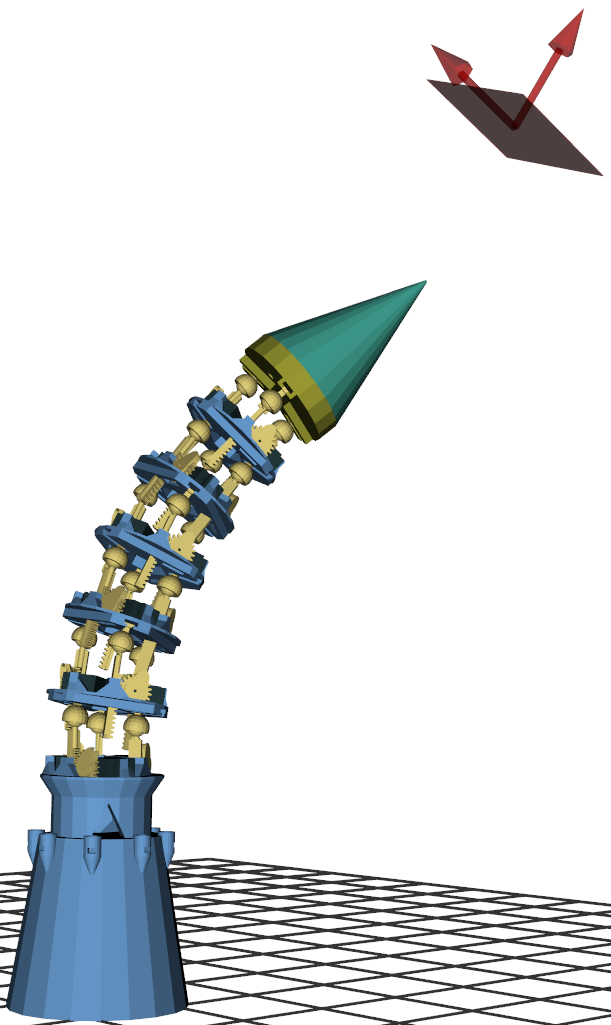}
        \elasticfigure{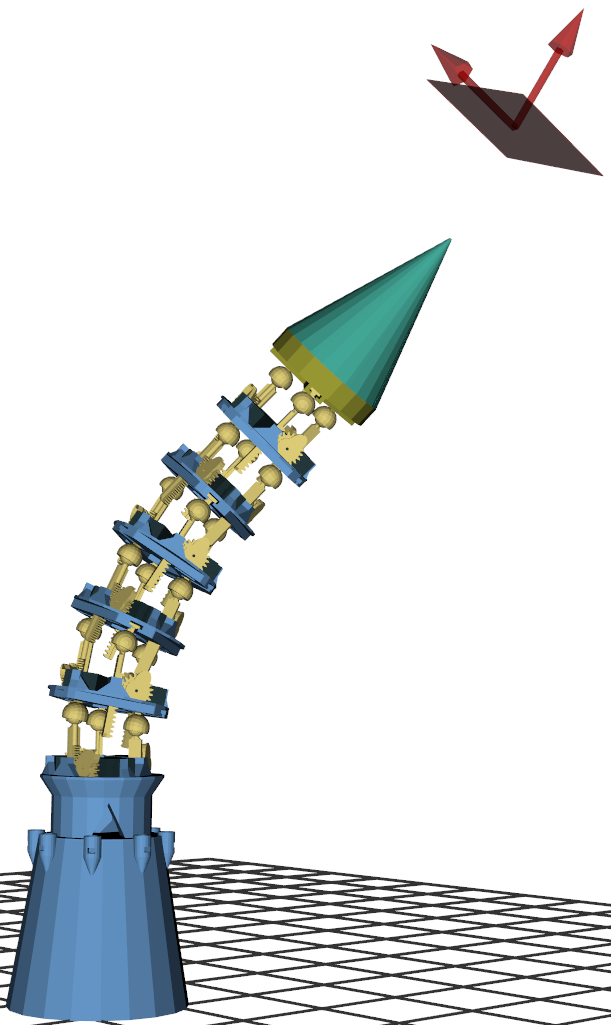}
        \elasticfigure{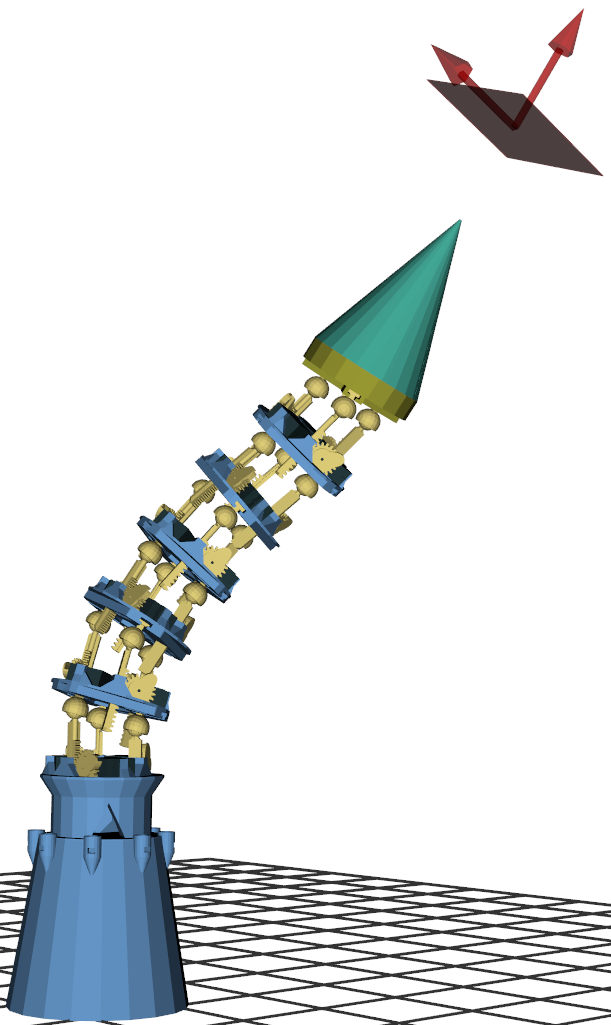}
        \elasticfigure{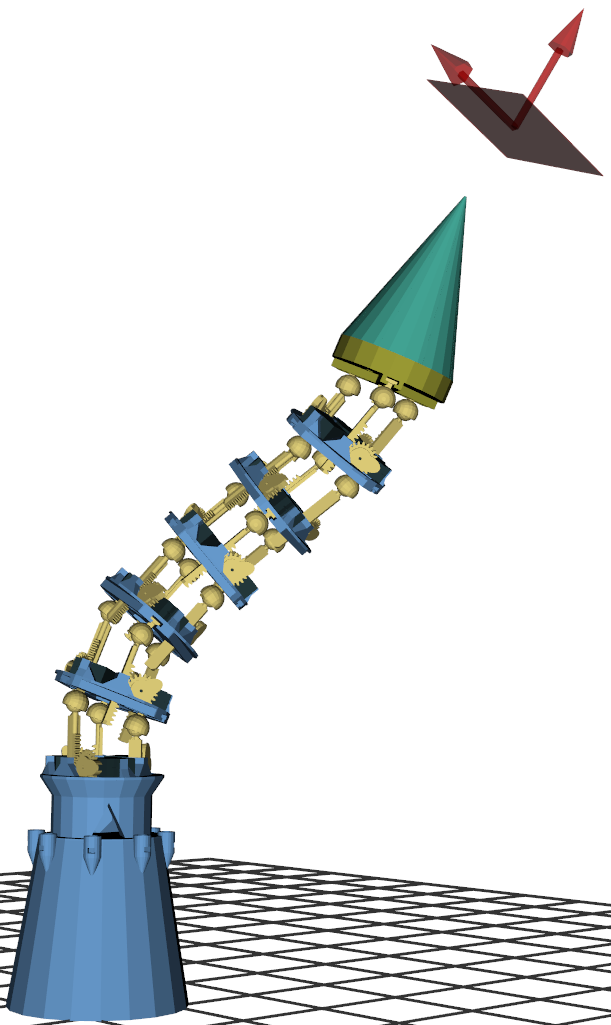}
        \elasticfigure{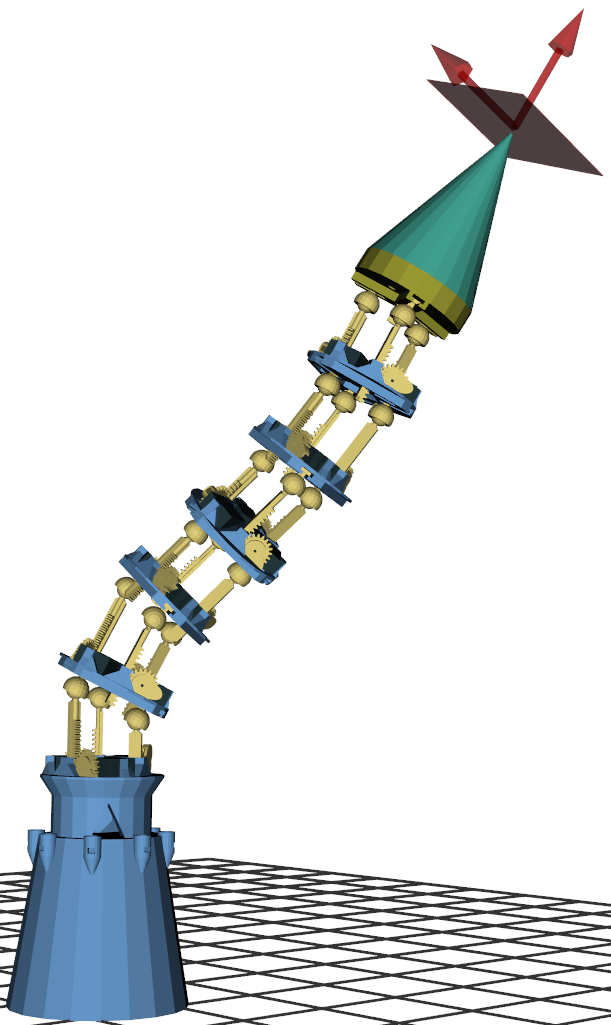}
    \end{elasticrow}
    \vskip\imagepaddingy
    \begin{elasticrow}[\imagepadding]
        \elasticfigure{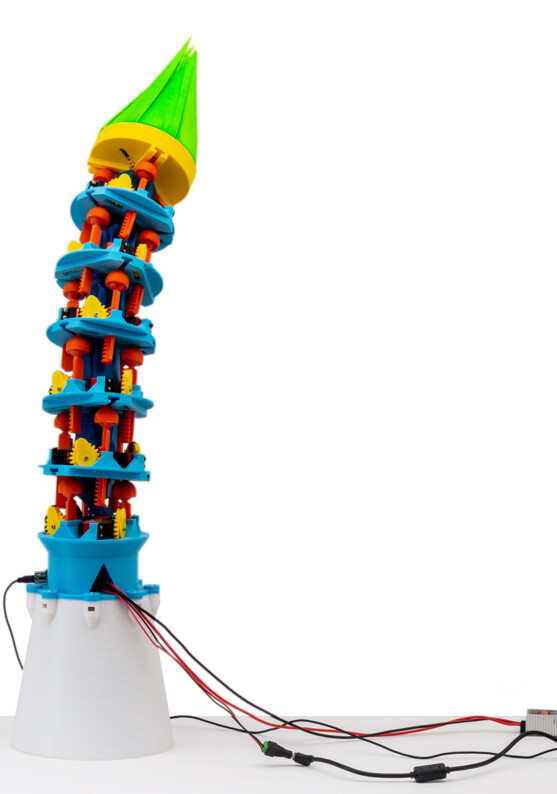}
        \elasticfigure{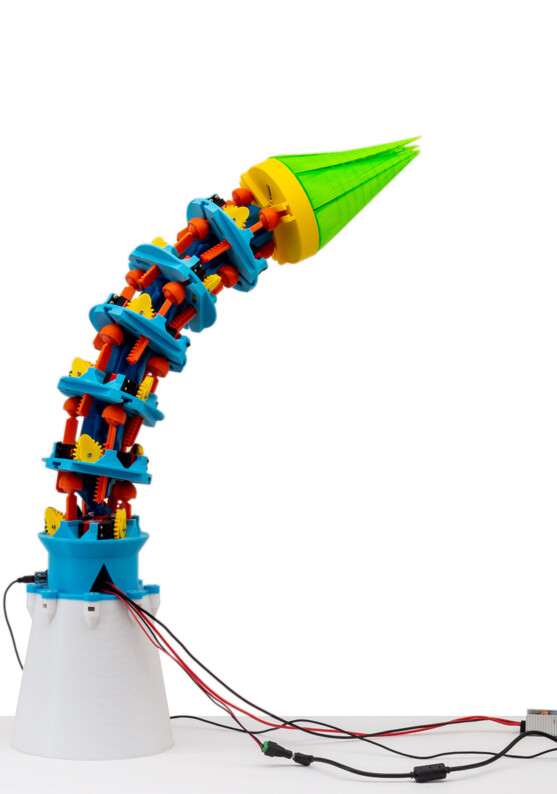}
        \elasticfigure{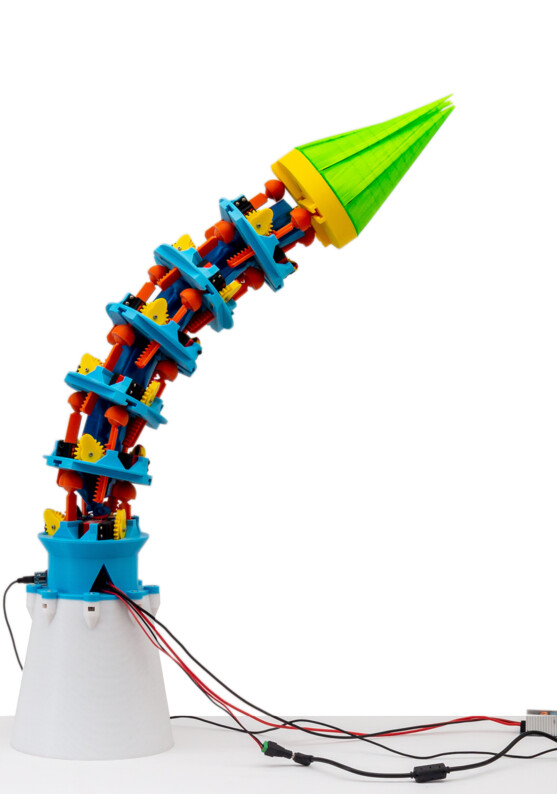}
        \elasticfigure{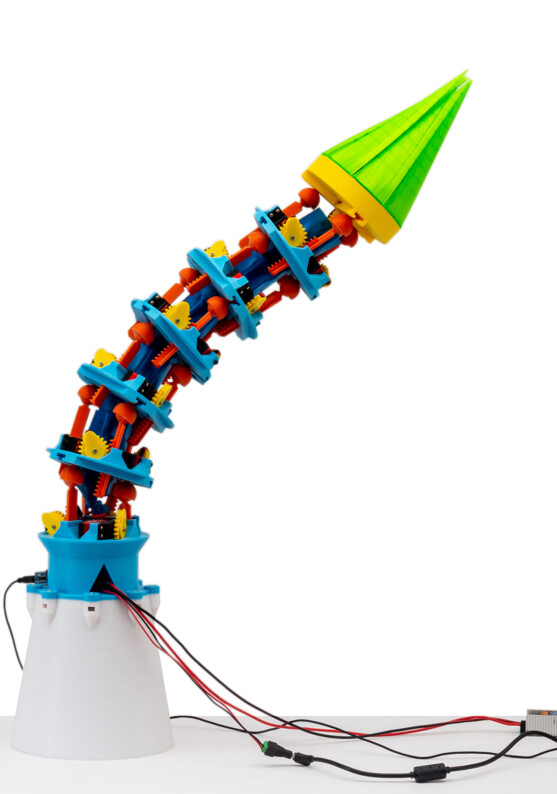}
        \elasticfigure{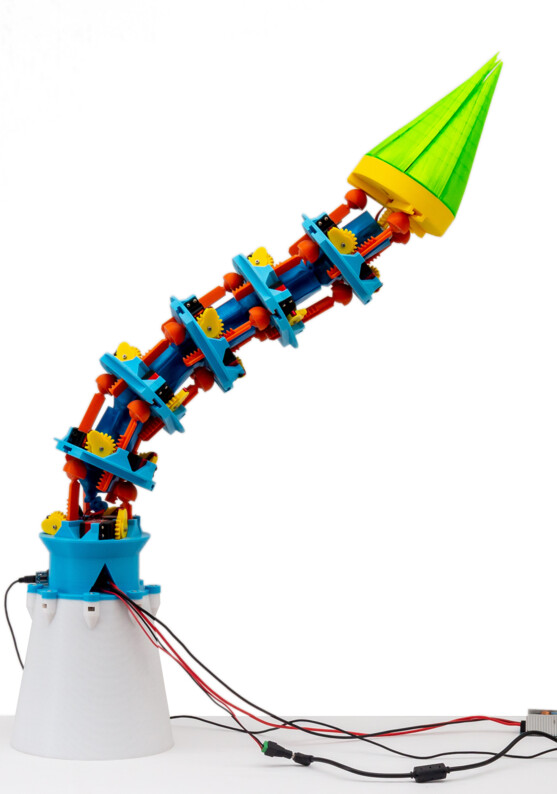}
        \elasticfigure{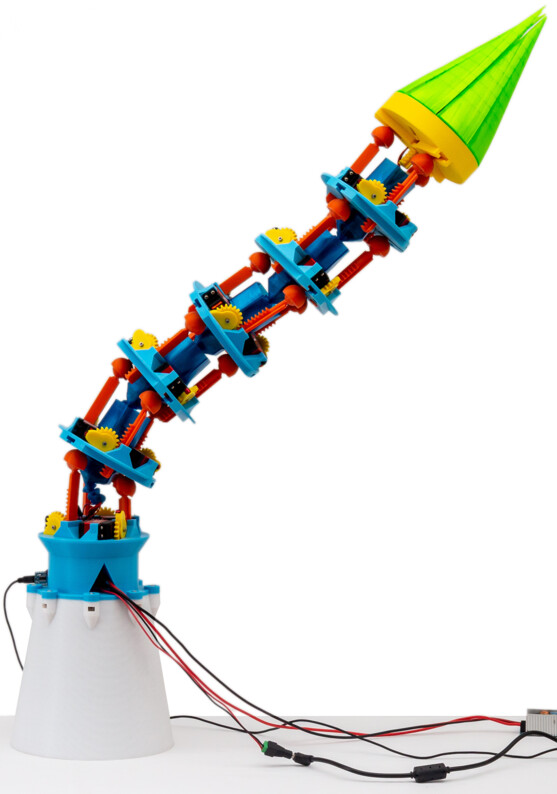}
    \end{elasticrow}

    \caption{Visualization of the inference process for the 3-geared robot. The top row shows the simulated robot approaching the red inference target with orientation visualized as arrows pointing in the relative x and z (up) direction. By transferring the inferred gear states to the real robot arm (shown in the bottom row), it then performs the same movement as in the simulation.}
    \label{fig:armv3}
\end{figure*}
\begin{figure*}[t!]
    \begin{elasticrow}[\imagepadding]
        \elasticfigure{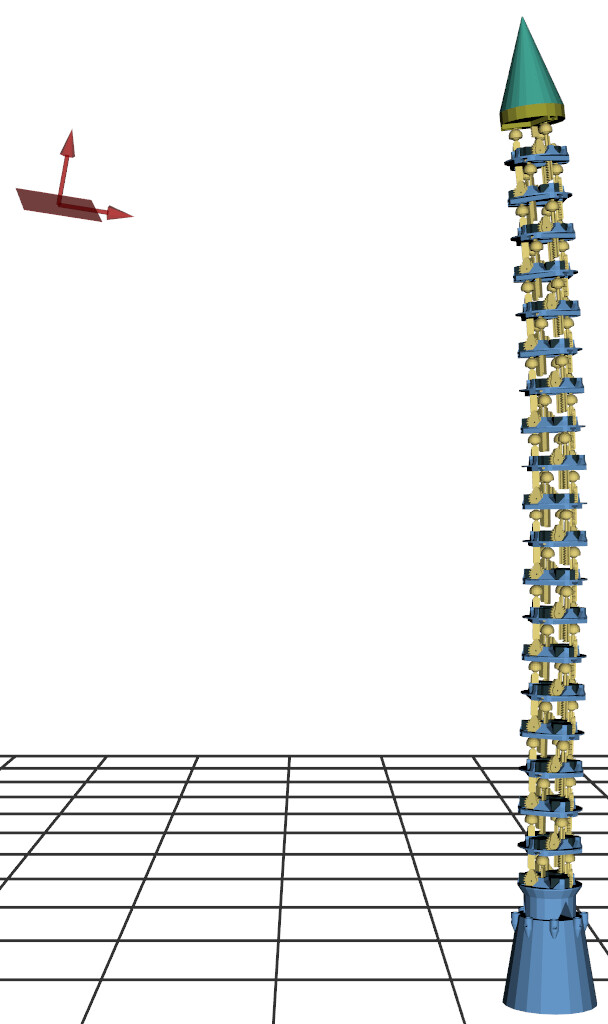}
        \elasticfigure{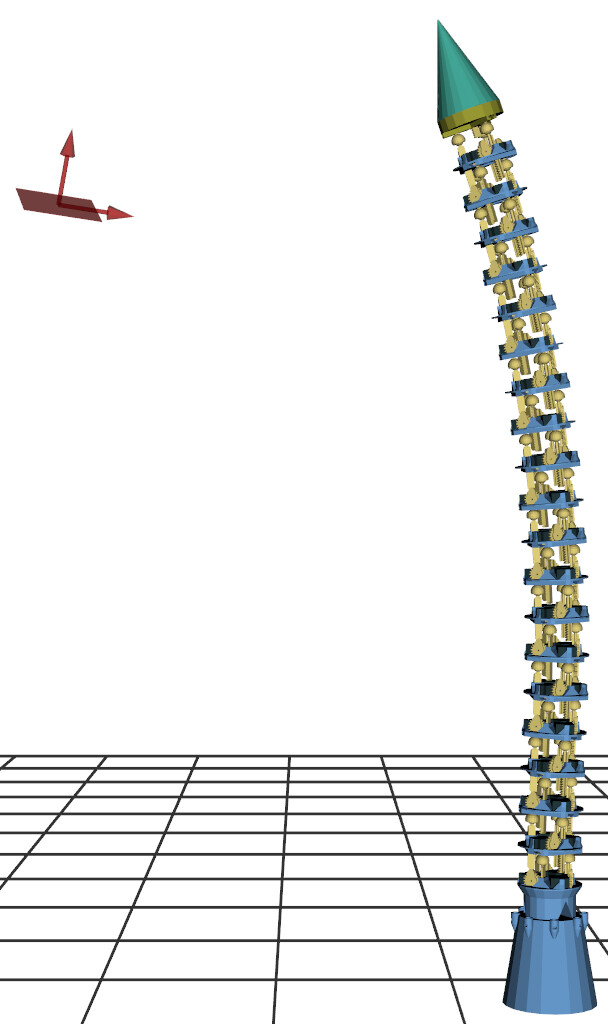}
        \elasticfigure{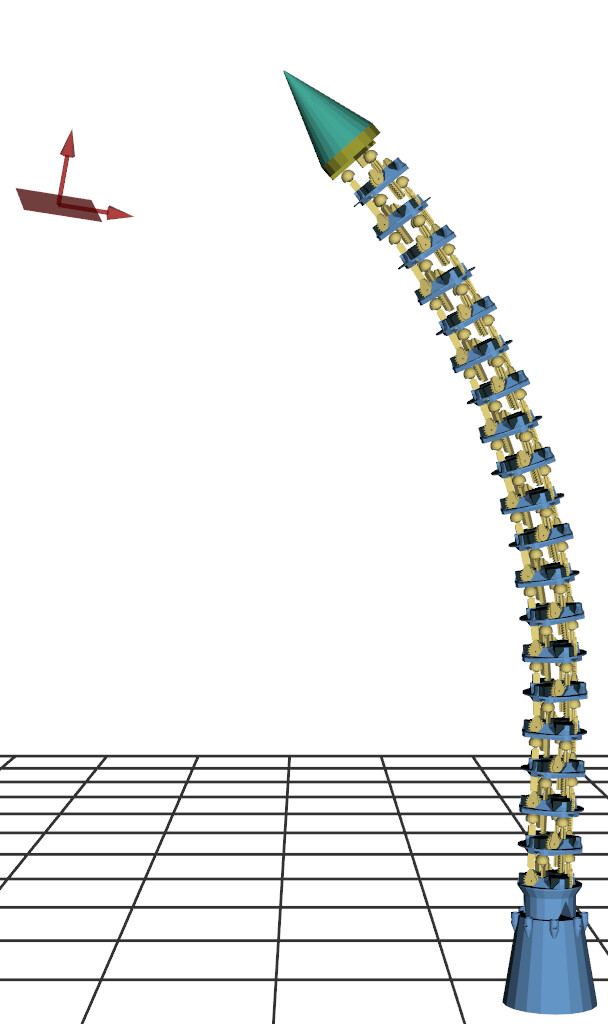}
        \elasticfigure{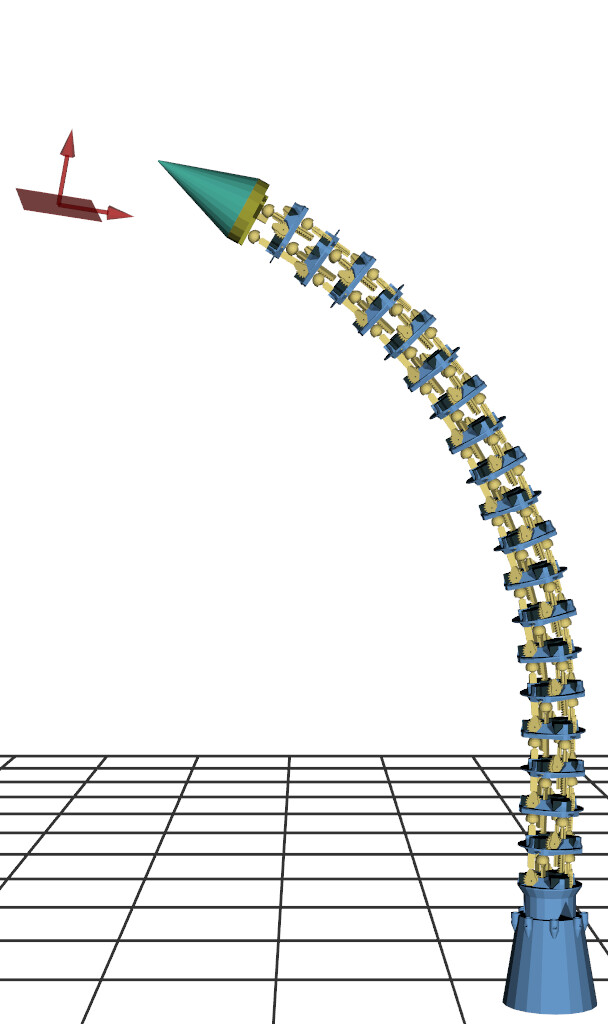}
        \elasticfigure{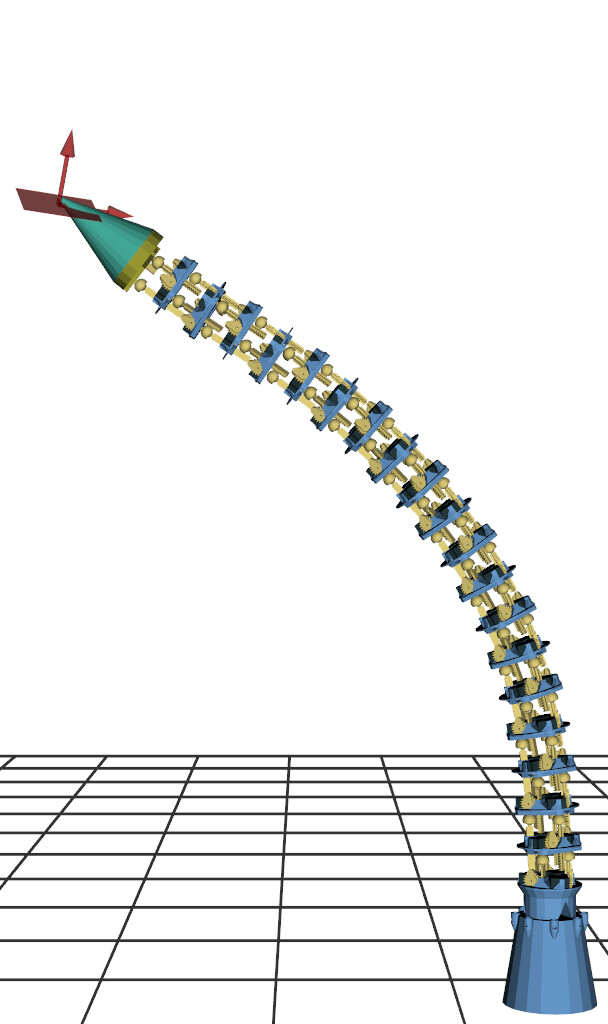}
        \elasticfigure{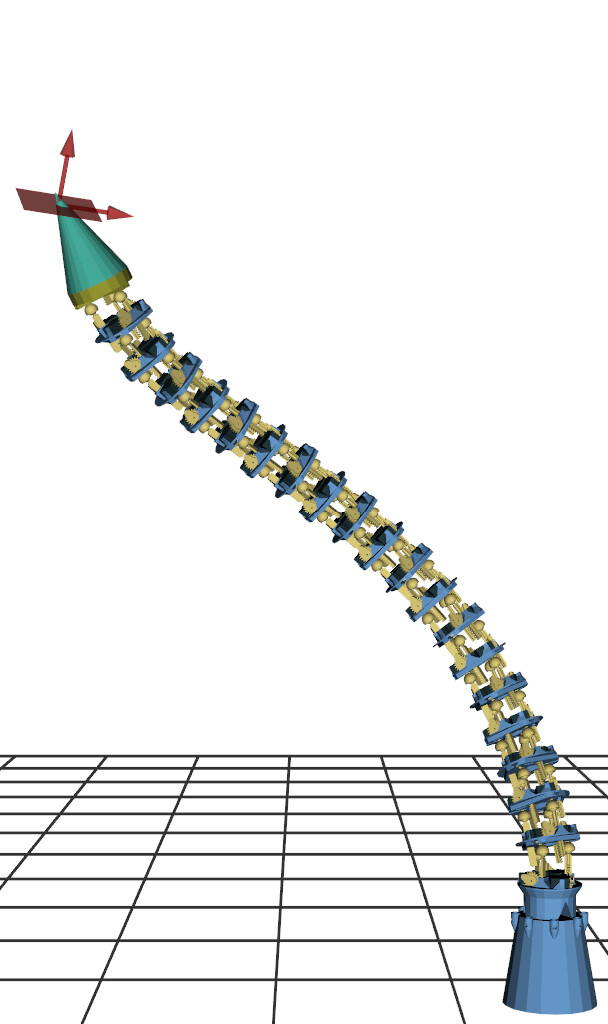}
        \elasticfigure{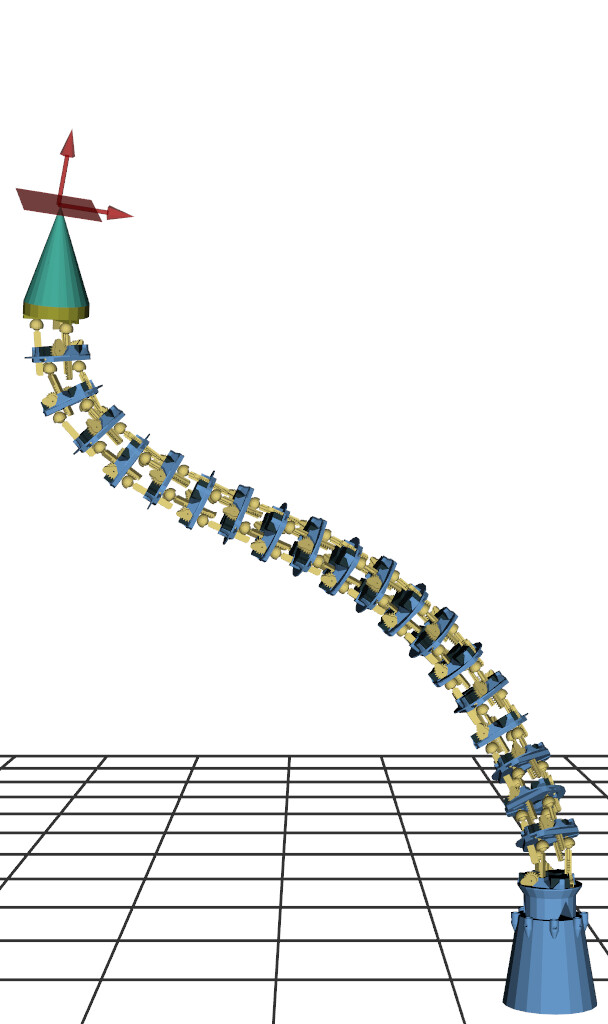}
    \end{elasticrow}
    \vskip\imagepaddingy
    \begin{elasticrow}[\imagepadding]
        \elasticfigure{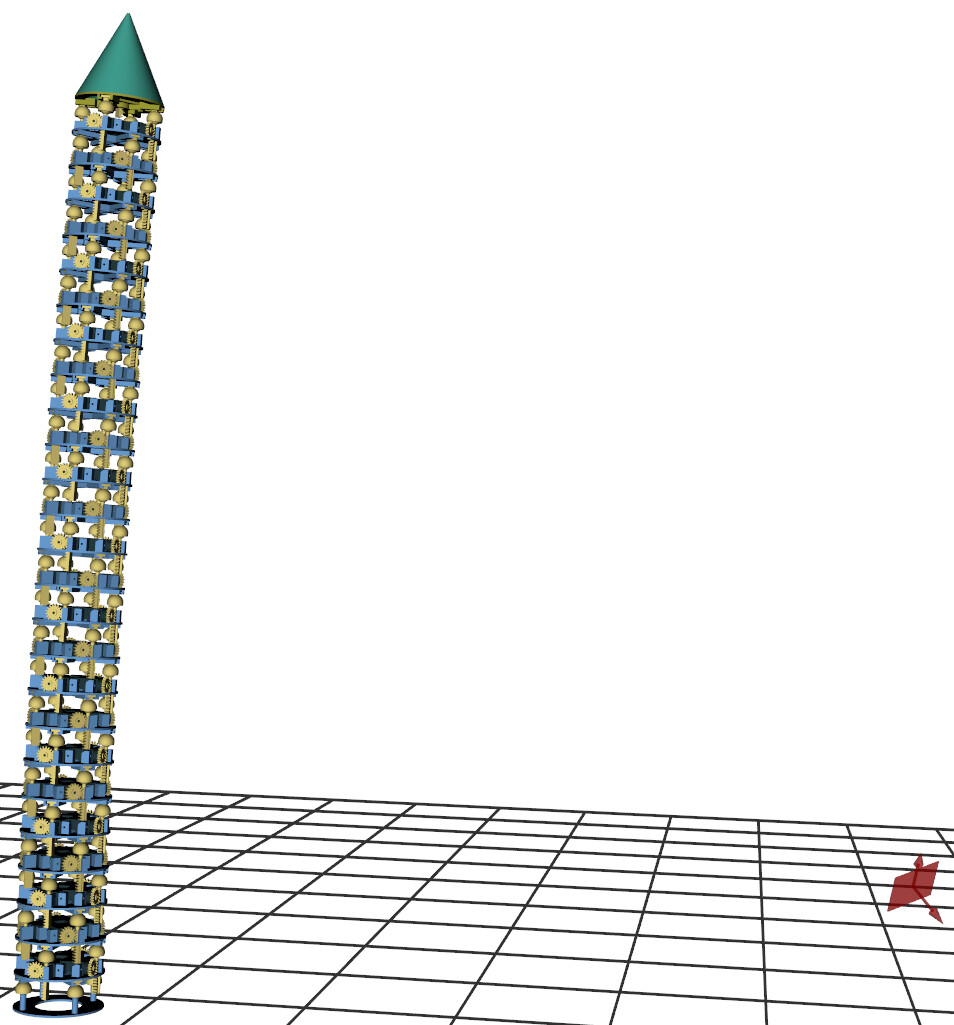}
        \elasticfigure{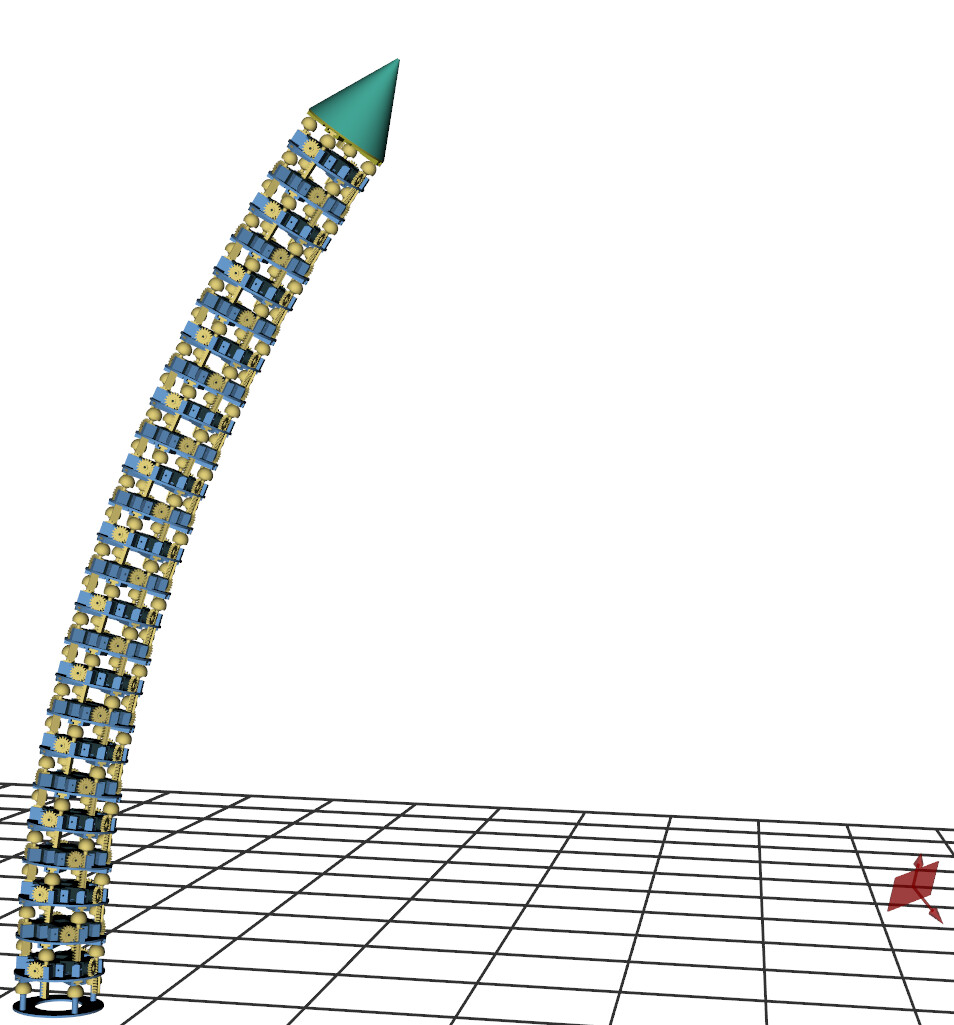}
        \elasticfigure{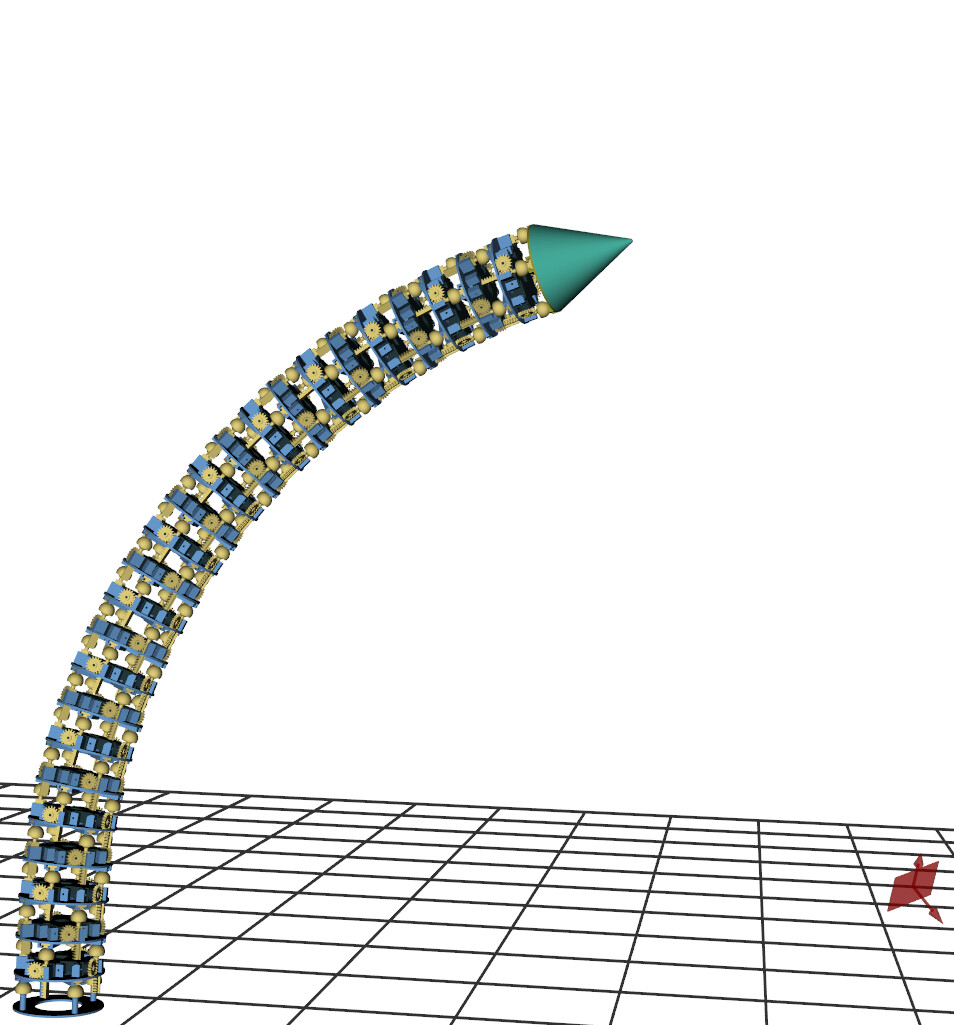}
        \elasticfigure{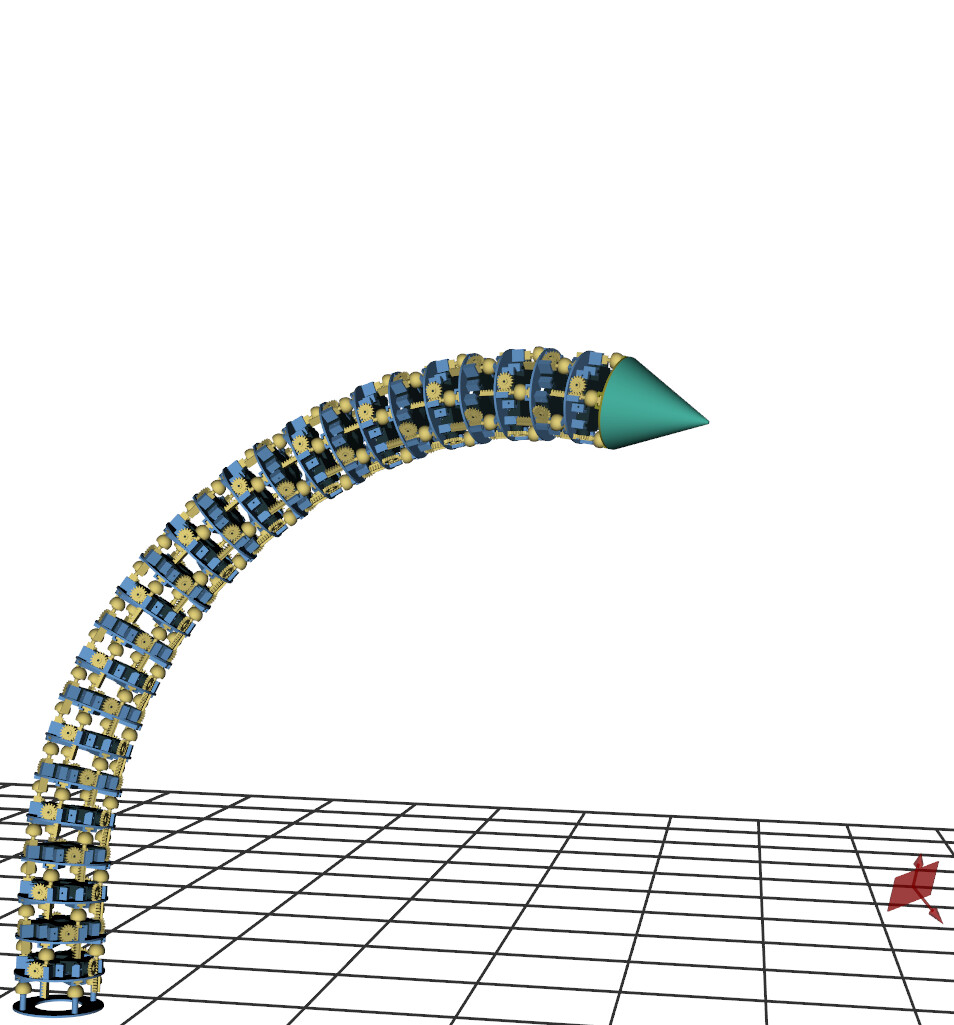}
        \elasticfigure{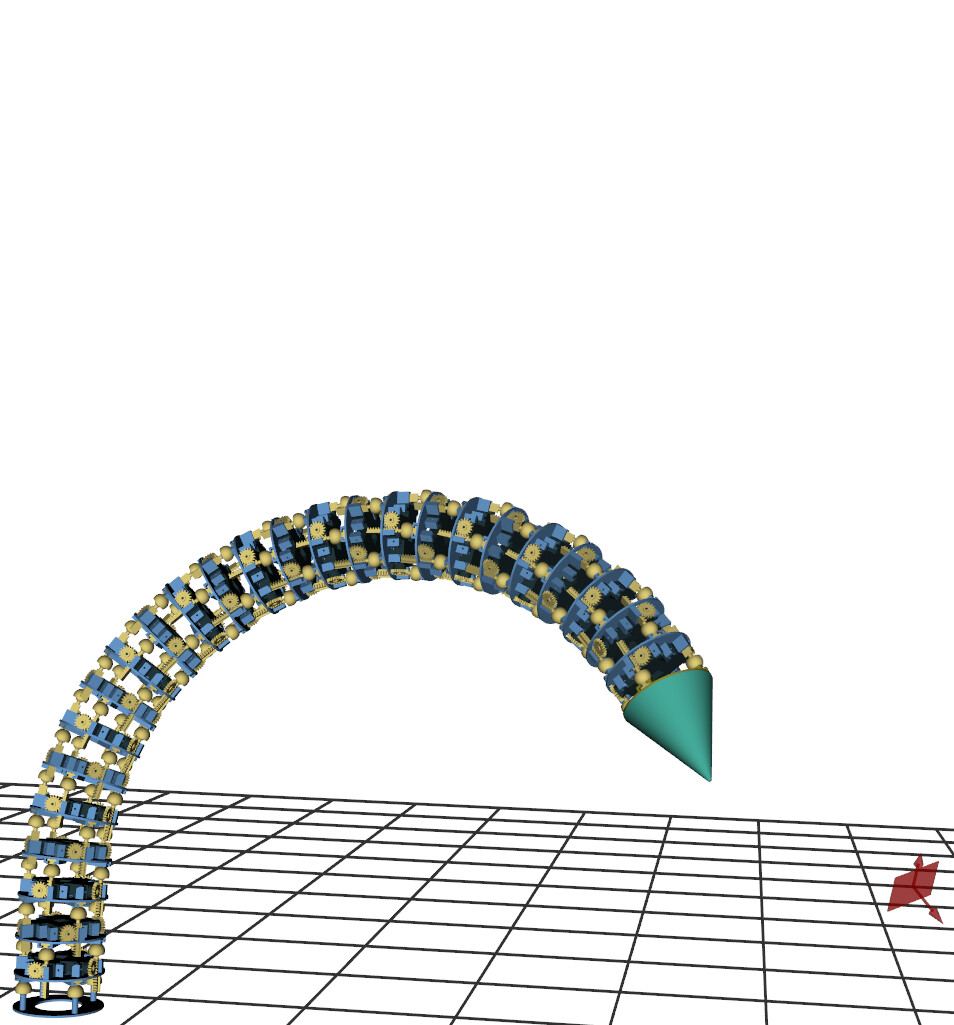}
        \elasticfigure{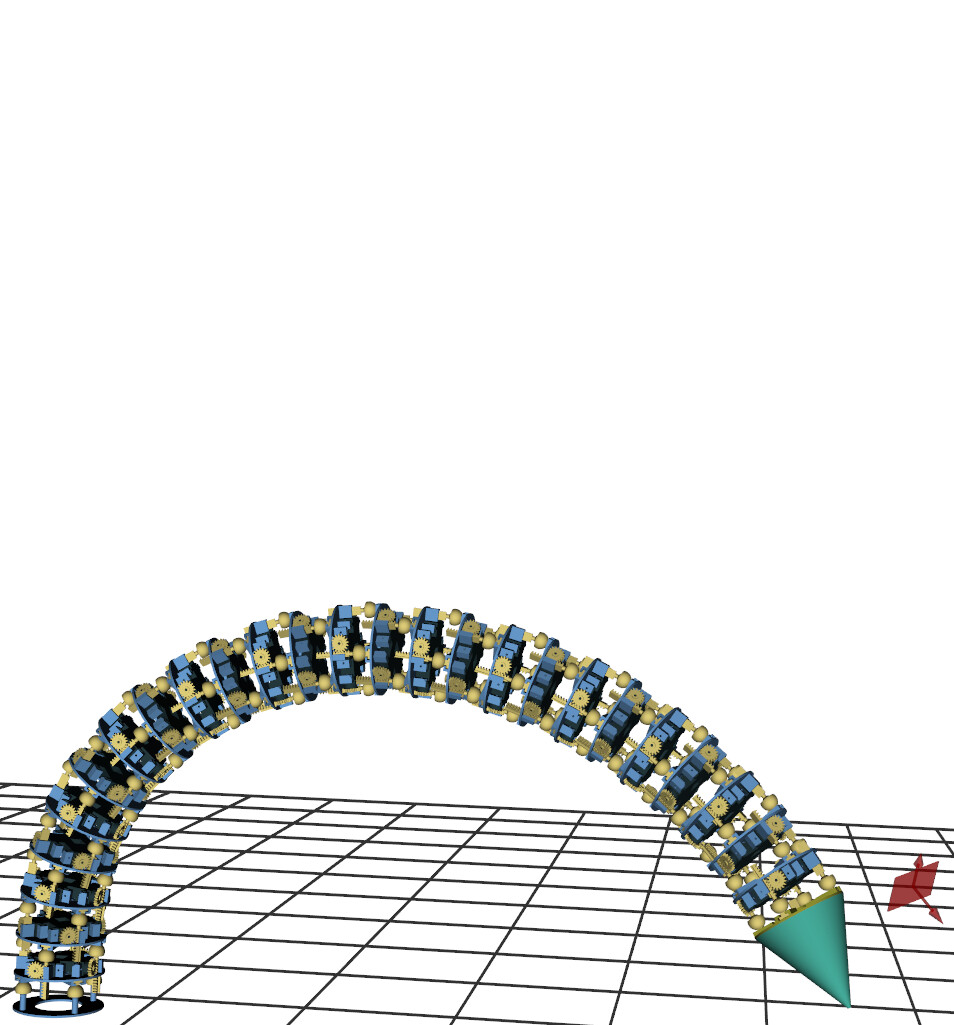}
        \elasticfigure{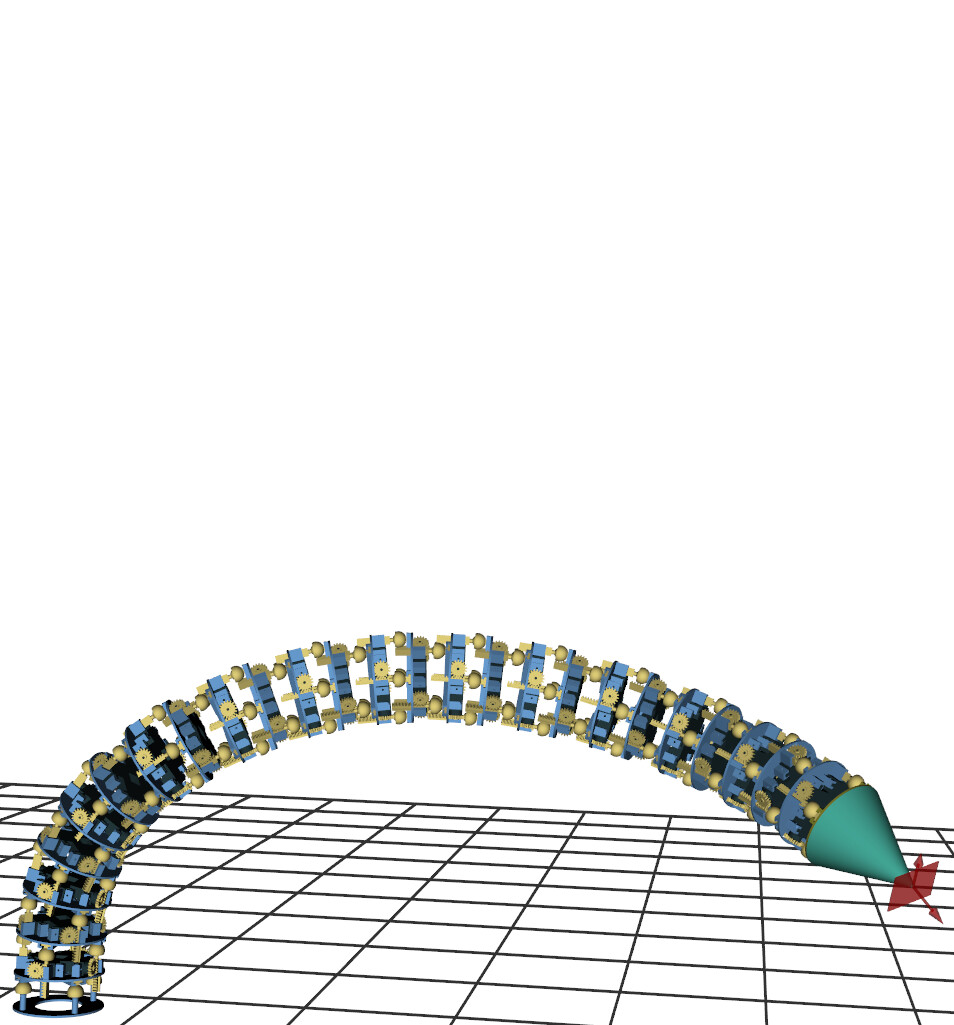}
    \end{elasticrow}
    
    \caption{Visualization of the inference process with long arms. \textbf{Top:} a 3-geared robot with 20 joints reaching for a target in the top left.
    \textbf{Bottom:} a 4-geared robot with 25 joints reaching for a target in the bottom right. While the position and orientation errors are simultaneously minimized in the inference process, one can see that in the beginning, the inference process is dominated by the position error. The orientation error only influences the inference process once the end-effector is close to the target.} 
    \label{fig:long_inference}
\end{figure*}

The first inference experiments were performed around the two actual 3D-printed robotic arms with 6 joints for the 3-geared design and 10 joints for the 4-geared design. 
\autoref{fig:armv3} shows an images sequences of the 3-geared arm driven by the SNN towards the target pose. The motion was actually executed on the real robot arm, whereas the simulation reflects the inference process simultaneously. \autoref{fig:long_inference} shows exemplary inference processes for longer arms, namely, a 20-joint robot using the 3-geared design in the top row and a 25-joint robot using the 4-geared design in the bottom row. We can see that the inference process in the beginning focuses more on the position, while the orientation is optimized later on when the end-effector is already very close to the target. Note that this behavior is not artificially introduced, since both position and orientation-error are minimized simultaneously, but it might be an effect of the normalization of positions, which are based on the mean inter-joint distance. This makes the influence of the position error greater than the orientation error up to the point where the end-effector is about one inter-joint distance away from the target. 

\begin{figure*}[t!]
    \centering
    \includegraphics[width=\linewidth]{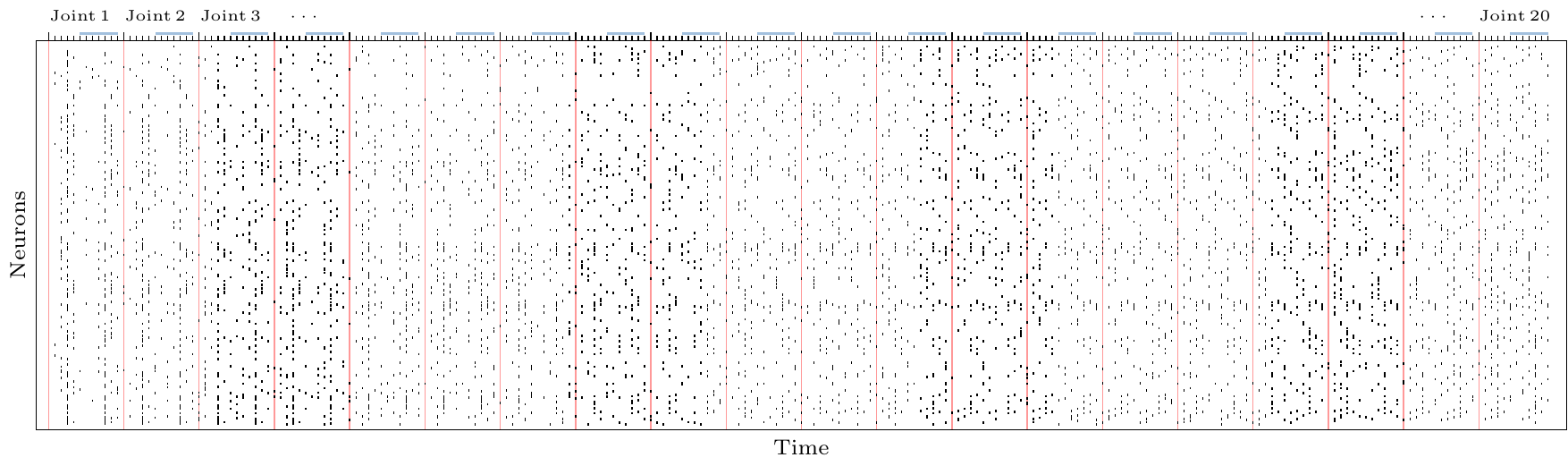}
    \caption{Exemplary depiction of the unfolding spike trains over 240 time steps within an LSNN with 512 hidden units while predicting the final posture of the 3-geared arm with 20 joints shown in \autoref{fig:long_inference} (top right). The vertical red lines indicate the first respective time step of the particular time periods (12 time steps) associated with the arm's joints. The blue marked time steps indicate the periods in which the clocking inputs are given and the output predictions are aggregated.}
    \label{fig:spike_train}
\end{figure*}

When observing the hidden spike trains of an LSNN during predicting a particular pose as shown in \autoref{fig:spike_train}, we can observe a rhythmic spiking behavior with two clear activity peaks within the time frame of a joint. However, the specific temporal pattern differs from joint to joint.

\begin{figure}[t!]
    \centering
    \includegraphics[]{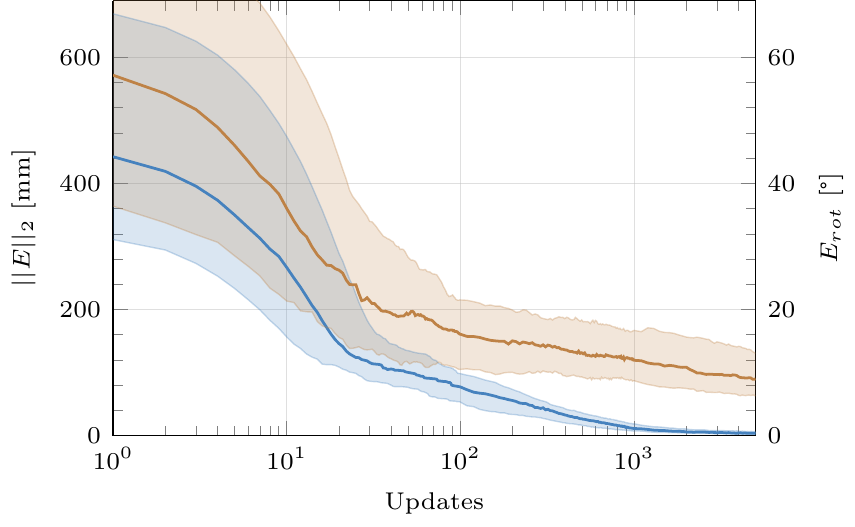}
    \caption{Inference position error \textbf{blue} vs orientation error \textbf{brown} for a 6-joint 3-geared robot. The forward model used 512 hidden neurons and the inference process was started with an initial learning-rate of $0.01$. The plot shows the median errors (dark lines) together with the upper and lower quantile as the filled areas. }
    \label{fig:inf_errors}
    \vspace{-0.5cm}
\end{figure}

\begin{figure}[t!]
    \centering
    \includegraphics[]{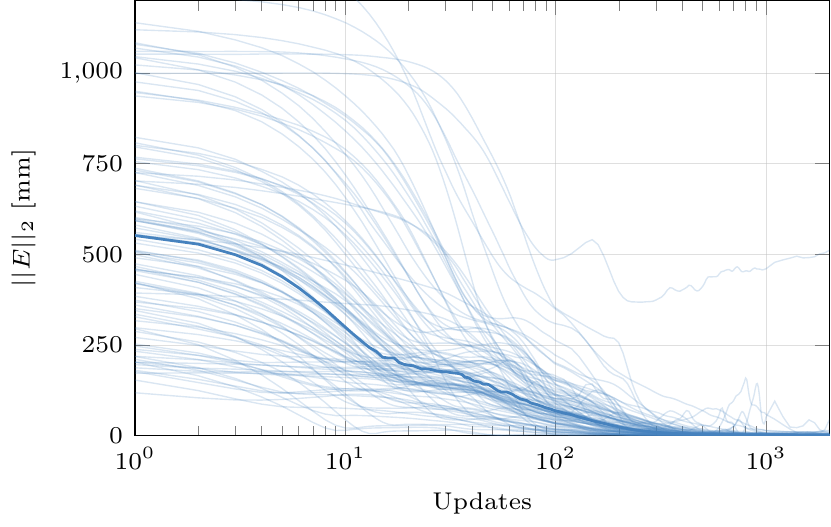}
    \caption{Inference process with a 25-joint 4-geared robot with over 100 random targets. The plot shows the Euclidean distance between the target and the end-effector (in millimeter) over time for different inference runs (\textbf{light blue}) and the median distance error over time (\textbf{blue}). The forward model was trained on an LSNN with 512 hidden neurons and the inference used an initial learning-rate of $0.01$. }
    \label{fig:spageti}
    \vspace{-0.2cm}
\end{figure}

The progression of position and orientation error across several independent runs can further be studied with   \autoref{fig:inf_errors}. While here the median position error already converged towards zero, the median orientation error did not go much below 10 degree during the 5,000 gradient updates of the inference process. For comparison \autoref{fig:spageti} shows the median distance error calculated over 100 independent inference runs augmented with the trajectories of the individual runs. Here the experiment shows that even for the simulated 25-joint (4-geared robot) the inference process consistently converges to a distance error in the sub centimeter regime and the median error even reaches $1$ mm.

\begin{figure}[t!]
    \centering
    \vspace{0.7cm}
    \includegraphics[]{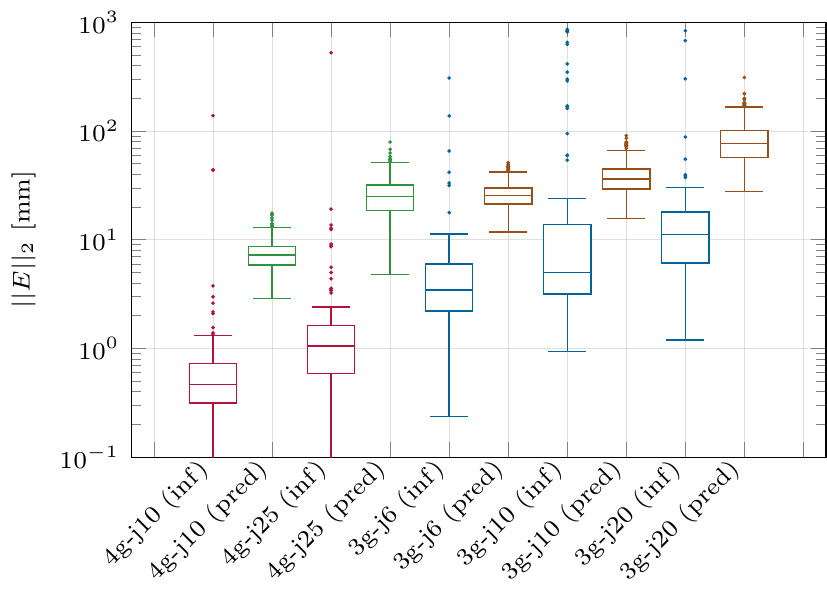}
    \caption{Comparison of inference distance errors and prediction errors for the different arm configurations: \textbf{red}: 4-gearerd robot inference errors, \textbf{green}: 4-geared robot prediction errors, \textbf{blue}: 3-geared robot inference errors, \textbf{brown}: 3-geared robot prediction errors. }
    \vspace{-0.7cm}
    \label{fig:correction}
\end{figure}

\autoref{fig:correction} provides an overview of the achieved control precision across multiple arms with target correction. The prediction errors for the forward models ranged from just below $1$ cm for the 10-joint 4-geared robot to around $10$ cm for the 20-joint 3-geared version. This difference in accuracy might be attributed to the greater size of the 3-geared robot, but also to the greater range of motion that each joint in the 3-geared robot has (40° vs 16°), which makes it harder for the network to learn the forward model. Despite the rather great difference in prediction accuracy, the prediction correction step manages to improve the inference position error consistently across the robotic arm models about one order of magnitude.

\begin{figure}[t!]
    \centering
    \includegraphics[]{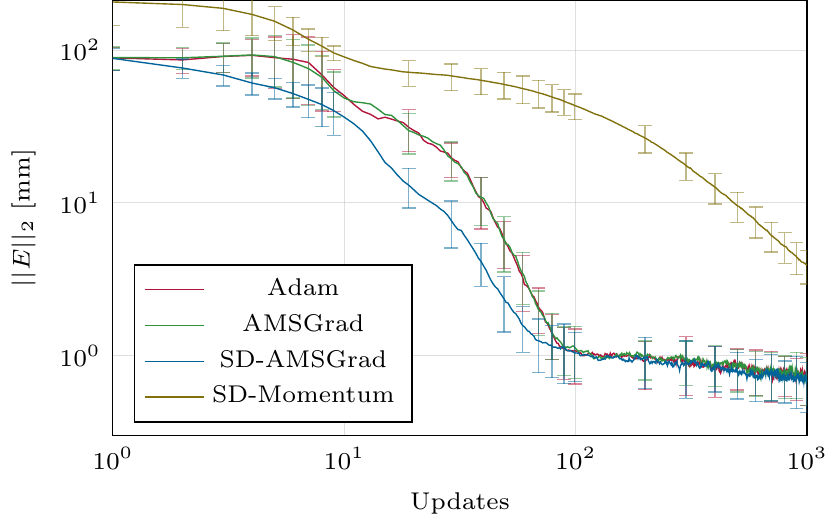}
    \caption{Comparison of different optimizers for the inference process, tested with a 10-joint 4-geared robot with a forward model using 128 hidden neurons and an initial learning-rate of $0.1$. The plot shows the median distance error together with the upper and lower quantile bars (calculated over 250 independent inference runs for each optimizer using 5 independent trained forward models)}
    \label{fig:optimizers}
\end{figure}

As an additional evaluation, \autoref{fig:optimizers} compares the median convergence behavior during motor inference for several optimizers. SD-Momentum has the disadvantage that it does not adaptively scale the gradients, both Adam and standard AMSGrad  suffer from an initial error increase. With the sign dampening mechanism, our SD-AMSGrad variant performs best. It provides quick, stable, and steady convergence and obtained the lowest inference error.

%


\section{CONCLUSION AND FUTURE WORK}

In this work we did not only show that it is possible to construct low-cost trunk-like robotic arms with basic 3D-printing equipment, but we also demonstrated how they can be controlled using the latest recurrent spiking neutral network architectures. Our method allows the precise goal-direction steering of these robots with near millimeter tolerance and it is able to handle even extraordinary complex robot arms with up to 75 articulated degrees of freedom. Although our current method of action inference depends on the backpropagation of prediction-error induced gradient signals, it might thus not yet be directly applicable on neuromorphic hardware, but it shows for the first time that spiking neural networks are capable of learning and controlling hyper redundant robotic arms this complex. The challenge ahead, however, is now to find suitable surrogate for BPTT in order to directly infer motor commands on neuromorphic hardware.

Future research could address the incorporation of radar-based distance sensors in order to implement integrative collision avoidance on real robot arms, which was previously done in second generation ANN-based simulation studies \cite{otte2018integrative}. This could also allow the robots to be used in soft robotic applications, where they might work together with humans in mutually beneficial interaction.

Another interesting research path would be the integration of this method into snake-like robots \cite{crespi2005amphibot,transeth2009survey} instead of stationary robotic arms, where they could be employed in search and rescue operations.
This could potentially allow the employment of even longer robots than those that are used as arms, and therefor would allow the utilization of the full potential of the LSNN networks. 



\bibliographystyle{IEEEtran}

\let\oldthebibliography\thebibliography
\let\endoldthebibliography\endthebibliography
\renewenvironment{thebibliography}[1]{
  \begin{oldthebibliography}{#1}
    \setlength{\itemsep}{0.05cm}
}
{
  \end{oldthebibliography}
}
\bibliography{2021-spikingarm}

\input{appendix}

\end{document}

%% file: appendix.tex
\clearpage

\appendices

\section{Additional parameters}

For the LSNN networks the LIF, ALIF, and readout decay factor was set to $\smash{\alpha = e^{-\frac{1}{20}}}$. For the adaptive threshold of ALIF neurons we used $\smash{\rho = e^{-\frac{1}{1200}}}$ and $\zeta = 0.03$. 
The base threshold for all neurons was set to $v_{thr} = 0.61$. 
The derivative dumping factor was set to $\gamma = 0.3$.
For the SD-AMSGrad algorithm (which was used of motor inference) we set $\beta_1 = 0.9$, $\beta_2 = 0.999$, and $\beta_3 = 0.9$.
For the target correction $\gamma_1 = 1$ and $\gamma_2 = 1$ was used.

\section{BPTT in LSNNs}\label{section:appendix:bptt}

In the following the equations to update the synaptic weights of LSNNs using BPTT are derived in detail. We assume an architecture with only one single recurrent hidden layer. Recall that for a given time step, the state of a neuron $j$ is denoted by the vector \smash{$\mathbf{s}_j^t$}. For LIF neurons this neuron state vector is one-dimensional and just 
contains the voltage \smash{$v_j^t$}, while for ALIF neurons it is two-dimensional and 
contains the voltage \smash{$v_j^t$} as well as the threshold adaption value \smash{$a_j^t$}:
\begin{equation}\label{eq:hidden_stae_lif_appendix}
    \vec{s}_{j,LIF}^t \defeq v_j^t
\end{equation}
\begin{equation}\label{eq:hidden_stae_alif_appendix}
    \vec{s}_{j,ALIF}^t \defeq 
    \begin{bmatrix}[1.4]
        v_j^t &
        a_j^t
    \end{bmatrix}^{\top}
\end{equation}

Also recall that we use a pseudo-derivative for the originally non-differentiable Heaviside activation function for the LIF and ALIF neurons:

\begin{equation}\label{eq:psudo_lif_appendix}
    \frac{\partial z_j^t}{\partial \vec{s}_{j,LIF}^t} \defeq h_{j,LIF}^t = \pseudoscale ~ \max \left( 0, \left\vert\frac{v_j^t - v_{thr}}{v_{thr}}\right\vert \right)
\end{equation}
\begin{equation}\label{eq:psudo_alif_appendix}
    \begin{split}
        \frac{\partial z_j^t}{\partial \vec{s}_{j,ALIF}^t} 
        &\defeq 
        \begin{bmatrix}[1.4]
            1 &
            - \zeta
        \end{bmatrix}^{\top}
        h_{j,ALIF}^t \\
        &=
        \begin{bmatrix}[1.4]
            1 &
            - \zeta
        \end{bmatrix}^{\top}
        \pseudoscale ~ \max \left( 0, \left\vert\frac{v_j^t - v_{j,thr}^t}{v_{thr}}\right\vert \right)
    \end{split}
\end{equation}

With this trick the model becomes differentiable and we can derive the error gradient for the loss $E$ with respect to all weights $w_{i,j}$ within the network.
%
%
Note that we leave $E$ unspecified here (we could, for instance, use the MSE summed over all involved time steps). 
\begin{equation}\label{eq:weight_appendix}
\frac{\partial E}{\partial w_{i,j}} = \sum_t \frac{\partial E}{\partial \mathbf{s}_j^t} \frac{\partial \mathbf{s}_j^t}{\partial w_{i,j}}
\end{equation}
The left term within the sum expresses how a change of the state of neuron $j$ at time step $t$ effects the change of the loss, which we define as:
\begin{equation}\label{eq:delta_def_appendix}
    \pmb{\delta}_j^t \defeq \frac{\partial E}{\partial \mathbf{s}_j^t}
\end{equation}
Applying the chain rule on this partial derivative gives us:
\begin{equation}\label{eq:delta_error_appendix}
    \frac{\partial E}{\partial \mathbf{s}_j^t}=\frac{\partial E}{\partial z_j^t} \frac{\partial z_j^t}{\partial \mathbf{s}_j^t} + 
    \frac{\partial E}{\partial \mathbf{s}_j^{t+1}} \frac{\partial \mathbf{s}_j^{t+1}}{\partial \mathbf{s}_j^t}
\end{equation}
It should be noted that the right sum term addresses specifically the neuron's internal state dependency over time excluding the gradient flow along the outer recurrencies (which is covered by the left sum term).

Further expanding the partial derivative of the error with respect to the output of neuron $j$ at time step $t$ results in:
\begin{equation}\label{eq:error_spike_appendix}
\frac{\partial E}{\partial z_j^t} = 
    \sum_k \frac{\partial E}{\partial \mathbf{s}_k^t} \frac{\partial \mathbf{s}_k^t}{\partial z_j^t} + 
    \sum_{j'} \frac{\partial E}{\partial \mathbf{s}_{j'}^{t+1}} \frac{\partial \mathbf{s}_{j'}^{t+1}}{\partial z_j^t} 
\end{equation}
where $k$ iterates over all successor neurons of $j$ (output neurons) and $j'$ iterates over all hidden neurons (including $j$). Note that the right sum term formulates the recursively nested error  dependency over time (BPTT).

All remaining terms in Equation~\eqref{eq:delta_error_appendix} and  Equation~\eqref{eq:error_spike_appendix} depend on the respective neuron model (LIF, ALIF or readout) and are further specified in the following.

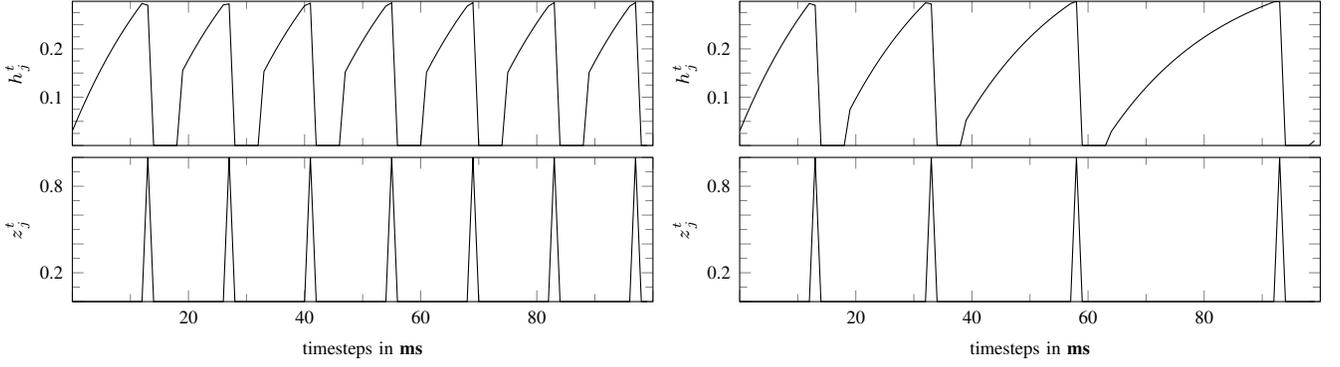
\begin{figure*}[t]
\begin{center}
\input{figures/pseudoshape/lif_derivatives.tex}
\input{figures/pseudoshape/alif_derivatives.tex}\\
\vspace{-0.2cm}
\input{figures/pseudoshape/lif_spikes.tex}
\input{figures/pseudoshape/alif_spikes.tex}
\end{center}
\caption{Comparison of LIF and ALIF neuron derivatives, with a derivative dumping factor of
           \smash{$\lambda = 0.3$}. \textbf{Left}: LIF neuron with a constant input 
          current of $0.1$ and a threshold of $v_{thr}=1.0$. \textbf{Right}: ALIF neuron with a constant 
          input current of $0.1$, a base threshold of $v_{thr}=1.0$, and a threshold increase of 
           $\zeta = 0.27$.}
           \label{fig:alif_lif_derivatives}
\end{figure*}


\subsection{Error Gradients for LIF Neurons}
As stated previously, the internal state of LIF neurons just contains the action potential $v_j^t$. Thus, deriving Equation~\eqref{eq:weight_appendix} with respect to the input weights results in:
\begin{equation}\label{eq:lifweightin_appendix}
\frac{\partial E}{\partial w_{i,j}^{in}} = \sum_t x_i^t \pmb{\delta}_j^t
\end{equation}
and deriving the equation with respect to the recurrent weights results in:
\begin{equation}\label{eq:lifweightrec_appendix}
\frac{\partial E}{\partial w_{j,j'}^{rec}} = \sum_t  z_{j}^{t} \pmb{\delta}_{j'}^{t+1}
\end{equation}

By substituting \smash{$\mathbf{s}_j^t$} with \smash{$v_j^t$} in Equation~ \eqref{eq:delta_error_appendix} and incorporating the pseudo-derivative from Equation~\eqref{eq:psudo_lif_appendix} we obtain:
\begin{equation}
\begin{split}
\pmb{\delta}_j^t &= \frac{\partial E}{\partial z_j^t} \frac{\partial z_j^t}{\partial v_j^t} + 
\frac{\partial E}{\partial v_j^{t+1}} \frac{\partial v_j^{t+1}}{\partial v_j^t}\\
           &= \frac{\partial E}{\partial z_j^t} h_j^t + \pmb{\delta}_j^{t+1} \alpha
\end{split}
\end{equation}

The partial derivative of the error with respect to the output spikes of a hidden neuron $j$ (left term in the previous equation) further decomposes into:
\begin{equation}\label{eq:error_spike_lif_appendix}
\begin{split}
\frac{\partial E}{\partial z_j^t} &= 
    \sum_k \frac{\partial E}{\partial \mathbf{s}_k^t} \frac{\partial \mathbf{s}_k^t}{\partial z_j^t} + 
    \sum_{j'} \frac{\partial E}{\partial v_{j'}^{t+1}} \frac{\partial v_{j'}^{t+1}}{\partial z_j^t}\\
    &= \sum_k \pmb{\delta}_k^t \frac{\partial \mathbf{s}_k^t}{\partial z_j^t} +
    \sum_{j'} \pmb{\delta}_{j'}^{t+1} \frac{\partial v_{j'}^{t+1}}{\partial z_j^t}
    \\
    &= \sum_k w_{j,k}^{out} \pmb{\delta}_{k}^t + \sum_{j'} w_{j,j'}^{rec} \pmb{\delta}_{j'}^{t+1} -
    \pmb{\delta}_j^{t+1} v_{thr}
\end{split}
\end{equation}
Note that the right most term here results from the internal recurrent dependency of the reset term in Equation~\eqref{eq:leaky_integrate_and_fire}. 

\subsection{Error Gradients for ALIF Neurons}
The internal state of ALIF neurons contains \smash{$v_j^t$} and \smash{$a_j^t$}. When we derive Equation~\eqref{eq:weight_appendix} with respect to the input weights, we eventually obtain:
\begin{equation}\label{eq:alifweightin_appendix}
\begin{split}
\frac{\partial E}{\partial w_{i,j}^{in}} &= \sum_t 
\begin{bmatrix}[1.4]
\frac{\partial E}{\partial v_j^t} &
\frac{\partial E}{\partial a_j^t}
\end{bmatrix}
\begin{bmatrix}[2]
\frac{\partial v_j^t}{\partial w_{i,j}^{in}} \\
\frac{\partial a_j^t}{\partial w_{i,j}^{in}}
\end{bmatrix} \\
&= \sum_t 
\begin{bmatrix}[1.4]
\delta_{j,v}^t & 
\delta_{j,a}^t
\end{bmatrix}
\begin{bmatrix}[1.4]
x_i^t \\
0
\end{bmatrix} \\
&= \sum_t \delta_{j,v}^t x_i^t
\end{split}
\end{equation}
and deriving the equation with respect to the recurrent weights gets us: 
\begin{equation}
\label{eq:alifweightrec_appendix}
\frac{\partial E}{\partial w_{j,j'}^{rec}} = \sum_t z_{j}^t \delta_{j',v}^{t+1} 
\end{equation}

The partial derivative reflecting the neuron's internal state dependency over time becomes a 2x2 matrix for ALIF neurons: 
\begin{equation}\label{eq:alif_state_derivative_bptt_appendix}
\frac{\partial \mathbf{s}_j^{t+1}}{\partial \mathbf{s}_j^t} = 
\begin{bmatrix}[2]
    \frac{\partial v_j^{t+1}}{\partial v_j^t} & \frac{\partial v_j^{t+1}}{\partial a_j^t} \\
    \frac{\partial a_j^{t+1}}{\partial v_j^t} & \frac{\partial a_j^{t+1}}{\partial a_j^t} 
\end{bmatrix} = 
\begin{bmatrix}[1.4]
    \alpha & 0 \\
    0 & \rho
\end{bmatrix}
\end{equation}  

When substituting this matrix in Equation~\ref{eq:delta_error_appendix} and including the pseudo-derivative from Equation~\eqref{eq:psudo_alif_appendix} we obtain:
\begin{equation}
\begin{split}
\pmb{\delta}_j^t &= \frac{\partial E}{\partial z_j^t} \frac{\partial z_j^t}{\partial \mathbf{s}_j^t} + 
\frac{\partial E}{\partial \mathbf{s}_j^{t+1}} \frac{\partial \mathbf{s}_j^{t+1}}{\partial \mathbf{s}_j^t} \\
    &= \frac{\partial E}{\partial z_j^t} 
    \begin{bmatrix}[1.4]
        h_j^t \\
        -h_j^t \zeta
    \end{bmatrix} 
+  
    \begin{bmatrix}[1.4]
        \alpha & 0 \\
        0 & \rho
    \end{bmatrix}
    \begin{bmatrix}[1.4]
       \delta_{j,v}^{t+1} \\
       \delta_{j,a}^{t+1}
    \end{bmatrix} \\
    &= \frac{\partial E}{\partial z_j^t} 
    \begin{bmatrix}[1.4]
        h_j^t \\
        -h_j^t \zeta
    \end{bmatrix} + 
    \begin{bmatrix}[1.4]
        \alpha \delta_{j,v}^{t+1} \\
        \rho \delta_{j,a}^{t+1} 
    \end{bmatrix}
\end{split}
\end{equation}

We can now further expand again the partial  derivative of the error with respect to the output spikes of neuron $j$ at time step $t$, which results in:
\begin{equation}
\begin{split}
\frac{\partial E}{\partial z_j^t} &= 
    \sum_k \frac{\partial E}{\partial \mathbf{s}_k^t} \frac{\partial \mathbf{s}_k^t}{\partial z_j^t} + 
    \sum_{j'} \frac{\partial E}{\partial \mathbf{s}_{j'}^{t+1}} \frac{\partial \mathbf{s}_{j'}^{t+1}}{\partial z_j^t} 
    \\
    &= \sum_k \pmb{\delta}_k^t w_{j,k}^{out} + \sum_{j'}
    \begin{bmatrix}[1.4]
        \delta_{j',v}^{t+1} &
        \delta_{j',a}^{t+1}
    \end{bmatrix} 
    \frac{\partial \mathbf{s}_{j'}^{t+1}}{\partial z_j^t}\\
    &= \sum_k \pmb{\delta}_k^t w_{j,k}^{out} + \sum_{j'} w_{j,j'}^{rec} \delta_{j',v}^{t+1}\\
    &~~~ -\delta_{j,v}^{t+1} v_{thr} + \delta_{j,a}^{t+1} 
\end{split}
\end{equation}

\subsection{Error Gradients for Output Synapses}

Before we can calculate the gradient for the output weights, we first have to derive error gradient with respect to the readout neurons:
\begin{equation}\label{eq:deltaout_appendix}
\begin{split}
\pmb{\delta}_k^t  &= \frac{\partial E}{\partial \mathbf{s}_k^t} + 
    \frac{\partial E}{\partial \mathbf{s}_k^{t+1}} \frac{\partial \mathbf{s}_k^{t+1}}{\partial \mathbf{s}_k^t} \\
    &= \frac{\partial E}{\partial \mathbf{s}_k^t}  + \pmb{\delta}_k^{t+1} \alpha
\end{split}
\end{equation}  
Using now, for instance, the following error function:
\begin{equation}
E = \frac{1}{2} \sum_{t}\sum_{k} \left(v_{k}^{t} - u_{k}^{t} \right)^{2}
\end{equation}
where \smash{$u_k^t$} denotes the target output for the $k$-th output neuron at time step $t$, we obtain: 
\begin{equation}\label{eq:deltaout_final_appendix}
\begin{split}
\pmb{\delta}_k^t  &=  (v_k^t - u_k^t)+ \pmb{\delta}_k^{t+1} \alpha
\end{split}
\end{equation}  

Finally, considering the gradient of the output weights and substituting $\pmb{\delta}_k^t$ results in:
\begin{equation}
\begin{split}
\frac{\partial E}{\partial w_{j,k}^{out}} &= \sum_t \frac{\partial E}{\partial \mathbf{s}_k^t} \frac{\partial \mathbf{s}_k^t}{\partial w_{j,k}^{out}} \\
&= \sum_t \left((v_k^t - u_k^t)+ \pmb{\delta}_k^{t+1} \alpha \right) z_j^{t} \\
&= \sum_t \sum_{t' \ge t} (v_k^{t'} - u_k^{t'}) \alpha^{t'-t} z_j^{t} 
\end{split}
\end{equation}

%% file: figures/pseudoshape/lif_derivatives.tex
\begin{tikzpicture}
    \begin{axis}[
      style={font=\scriptsize},
      width=9.3cm,
      height=3.5cm,
      xmin=0,
      xmax=100,
      ymax=0.298443,
      ymin=0,
      minor y tick num=1,
      minor x tick num=1,
      legend pos=south east,
      ylabel={\smash{$h_j^t$}},
      try min ticks=5,
      yticklabels={,,,0.1,,0.2,},
      xticklabels={,,},
    ]
        \addplot[mark=none,black] plot coordinates {
(0.000000, 0.030000)
(1.000000, 0.058537)
(2.000000, 0.085682)
(3.000000, 0.111503)
(4.000000, 0.136065)
(5.000000, 0.159429)
(6.000000, 0.181654)
(7.000000, 0.202794)
(8.000000, 0.222904)
(9.000000, 0.242033)
(10.000000, 0.260229)
(11.000000, 0.277537)
(12.000000, 0.294002)
(13.000000, 0.290337)
(14.000000, 0.000000)
(15.000000, 0.000000)
(16.000000, 0.000000)
(17.000000, 0.000000)
(18.000000, 0.000000)
(19.000000, 0.155193)
(20.000000, 0.177624)
(21.000000, 0.198961)
(22.000000, 0.219258)
(23.000000, 0.238564)
(24.000000, 0.256930)
(25.000000, 0.274399)
(26.000000, 0.291016)
(27.000000, 0.293177)
(28.000000, 0.000000)
(29.000000, 0.000000)
(30.000000, 0.000000)
(31.000000, 0.000000)
(32.000000, 0.000000)
(33.000000, 0.153089)
(34.000000, 0.175623)
(35.000000, 0.197058)
(36.000000, 0.217447)
(37.000000, 0.236842)
(38.000000, 0.255291)
(39.000000, 0.272840)
(40.000000, 0.289534)
(41.000000, 0.294587)
(42.000000, 0.000000)
(43.000000, 0.000000)
(44.000000, 0.000000)
(45.000000, 0.000000)
(46.000000, 0.000000)
(47.000000, 0.152045)
(48.000000, 0.174629)
(49.000000, 0.196113)
(50.000000, 0.216548)
(51.000000, 0.235987)
(52.000000, 0.254478)
(53.000000, 0.272067)
(54.000000, 0.288798)
(55.000000, 0.295287)
(56.000000, 0.000000)
(57.000000, 0.000000)
(58.000000, 0.000000)
(59.000000, 0.000000)
(60.000000, 0.000000)
(61.000000, 0.151526)
(62.000000, 0.174136)
(63.000000, 0.195643)
(64.000000, 0.216102)
(65.000000, 0.235562)
(66.000000, 0.254074)
(67.000000, 0.271682)
(68.000000, 0.288432)
(69.000000, 0.295635)
(70.000000, 0.000000)
(71.000000, 0.000000)
(72.000000, 0.000000)
(73.000000, 0.000000)
(74.000000, 0.000000)
(75.000000, 0.151268)
(76.000000, 0.173891)
(77.000000, 0.195410)
(78.000000, 0.215880)
(79.000000, 0.235351)
(80.000000, 0.253873)
(81.000000, 0.271491)
(82.000000, 0.288251)
(83.000000, 0.295807)
(84.000000, 0.000000)
(85.000000, 0.000000)
(86.000000, 0.000000)
(87.000000, 0.000000)
(88.000000, 0.000000)
(89.000000, 0.151140)
(90.000000, 0.173769)
(91.000000, 0.195294)
(92.000000, 0.215770)
(93.000000, 0.235246)
(94.000000, 0.253773)
(95.000000, 0.271397)
(96.000000, 0.288161)
(97.000000, 0.295893)
(98.000000, 0.000000)
(99.000000, 0.000000)
 };
    \end{axis}
\end{tikzpicture}

%% file: figures/pseudoshape/alif_derivatives.tex
\begin{tikzpicture}
    \begin{axis}[
      style={font=\scriptsize},
      width=9.3cm,
      height=3.5cm,
      xmin=0,
      xmax=100,
      ymax=0.298443,
      ymin=0,
      minor y tick num=1,
      minor x tick num=1,
      legend pos=south east,
      ylabel={\smash{$h_j^t$}},
      try min ticks=5,
      yticklabels={,,,0.1,,0.2,},
      xticklabels={,,},
    ]
        \addplot[mark=none,black] plot coordinates {
(0.000000, 0.030000)
(1.000000, 0.058537)
(2.000000, 0.085682)
(3.000000, 0.111503)
(4.000000, 0.136065)
(5.000000, 0.159429)
(6.000000, 0.181654)
(7.000000, 0.202794)
(8.000000, 0.222904)
(9.000000, 0.242033)
(10.000000, 0.260229)
(11.000000, 0.277537)
(12.000000, 0.294002)
(13.000000, 0.290337)
(14.000000, 0.000000)
(15.000000, 0.000000)
(16.000000, 0.000000)
(17.000000, 0.000000)
(18.000000, 0.000000)
(19.000000, 0.074530)
(20.000000, 0.097028)
(21.000000, 0.118432)
(22.000000, 0.138796)
(23.000000, 0.158170)
(24.000000, 0.176602)
(25.000000, 0.194138)
(26.000000, 0.210822)
(27.000000, 0.226696)
(28.000000, 0.241799)
(29.000000, 0.256168)
(30.000000, 0.269840)
(31.000000, 0.282848)
(32.000000, 0.295225)
(33.000000, 0.292998)
(34.000000, 0.000000)
(35.000000, 0.000000)
(36.000000, 0.000000)
(37.000000, 0.000000)
(38.000000, 0.000000)
(39.000000, 0.052292)
(40.000000, 0.072072)
(41.000000, 0.090894)
(42.000000, 0.108804)
(43.000000, 0.125847)
(44.000000, 0.142065)
(45.000000, 0.157499)
(46.000000, 0.172186)
(47.000000, 0.186163)
(48.000000, 0.199465)
(49.000000, 0.212125)
(50.000000, 0.224173)
(51.000000, 0.235641)
(52.000000, 0.246555)
(53.000000, 0.256943)
(54.000000, 0.266831)
(55.000000, 0.276243)
(56.000000, 0.285203)
(57.000000, 0.293731)
(58.000000, 0.298150)
(59.000000, 0.000000)
(60.000000, 0.000000)
(61.000000, 0.000000)
(62.000000, 0.000000)
(63.000000, 0.000000)
(64.000000, 0.028712)
(65.000000, 0.045933)
(66.000000, 0.062324)
(67.000000, 0.077925)
(68.000000, 0.092775)
(69.000000, 0.106909)
(70.000000, 0.120364)
(71.000000, 0.133172)
(72.000000, 0.145365)
(73.000000, 0.156973)
(74.000000, 0.168024)
(75.000000, 0.178546)
(76.000000, 0.188563)
(77.000000, 0.198102)
(78.000000, 0.207184)
(79.000000, 0.215834)
(80.000000, 0.224070)
(81.000000, 0.231915)
(82.000000, 0.239386)
(83.000000, 0.246502)
(84.000000, 0.253280)
(85.000000, 0.259737)
(86.000000, 0.265889)
(87.000000, 0.271749)
(88.000000, 0.277334)
(89.000000, 0.282655)
(90.000000, 0.287726)
(91.000000, 0.292559)
(92.000000, 0.297165)
(93.000000, 0.298443)
(94.000000, 0.000000)
(95.000000, 0.000000)
(96.000000, 0.000000)
(97.000000, 0.000000)
(98.000000, 0.000000)
(99.000000, 0.009630)
 };
    \end{axis}
\end{tikzpicture}

%% file: figures/pseudoshape/lif_spikes.tex
\begin{tikzpicture}
    \begin{axis}[
      style={font=\scriptsize},
      width=9.3cm,
      height=3.5cm,
      xmin=0,
      xmax=100,
      ymax=1,
      ymin=0,
      minor y tick num=1,
      minor x tick num=1,
      legend pos=south east,
      xlabel={timesteps in \textbf{ms}},
      xticklabels={,,20,40,60,80,},
      yticklabels={,,0.2,,,0.8,},
      try min ticks=5,
      ylabel={\smash{$z_j^t$}},
    ]
        \addplot[mark=none,black] plot coordinates {
(0.000000, 0.000000)
(1.000000, 0.000000)
(2.000000, 0.000000)
(3.000000, 0.000000)
(4.000000, 0.000000)
(5.000000, 0.000000)
(6.000000, 0.000000)
(7.000000, 0.000000)
(8.000000, 0.000000)
(9.000000, 0.000000)
(10.000000, 0.000000)
(11.000000, 0.000000)
(12.000000, 0.000000)
(13.000000, 1.000000)
(14.000000, 0.000000)
(15.000000, 0.000000)
(16.000000, 0.000000)
(17.000000, 0.000000)
(18.000000, 0.000000)
(19.000000, 0.000000)
(20.000000, 0.000000)
(21.000000, 0.000000)
(22.000000, 0.000000)
(23.000000, 0.000000)
(24.000000, 0.000000)
(25.000000, 0.000000)
(26.000000, 0.000000)
(27.000000, 1.000000)
(28.000000, 0.000000)
(29.000000, 0.000000)
(30.000000, 0.000000)
(31.000000, 0.000000)
(32.000000, 0.000000)
(33.000000, 0.000000)
(34.000000, 0.000000)
(35.000000, 0.000000)
(36.000000, 0.000000)
(37.000000, 0.000000)
(38.000000, 0.000000)
(39.000000, 0.000000)
(40.000000, 0.000000)
(41.000000, 1.000000)
(42.000000, 0.000000)
(43.000000, 0.000000)
(44.000000, 0.000000)
(45.000000, 0.000000)
(46.000000, 0.000000)
(47.000000, 0.000000)
(48.000000, 0.000000)
(49.000000, 0.000000)
(50.000000, 0.000000)
(51.000000, 0.000000)
(52.000000, 0.000000)
(53.000000, 0.000000)
(54.000000, 0.000000)
(55.000000, 1.000000)
(56.000000, 0.000000)
(57.000000, 0.000000)
(58.000000, 0.000000)
(59.000000, 0.000000)
(60.000000, 0.000000)
(61.000000, 0.000000)
(62.000000, 0.000000)
(63.000000, 0.000000)
(64.000000, 0.000000)
(65.000000, 0.000000)
(66.000000, 0.000000)
(67.000000, 0.000000)
(68.000000, 0.000000)
(69.000000, 1.000000)
(70.000000, 0.000000)
(71.000000, 0.000000)
(72.000000, 0.000000)
(73.000000, 0.000000)
(74.000000, 0.000000)
(75.000000, 0.000000)
(76.000000, 0.000000)
(77.000000, 0.000000)
(78.000000, 0.000000)
(79.000000, 0.000000)
(80.000000, 0.000000)
(81.000000, 0.000000)
(82.000000, 0.000000)
(83.000000, 1.000000)
(84.000000, 0.000000)
(85.000000, 0.000000)
(86.000000, 0.000000)
(87.000000, 0.000000)
(88.000000, 0.000000)
(89.000000, 0.000000)
(90.000000, 0.000000)
(91.000000, 0.000000)
(92.000000, 0.000000)
(93.000000, 0.000000)
(94.000000, 0.000000)
(95.000000, 0.000000)
(96.000000, 0.000000)
(97.000000, 1.000000)
(98.000000, 0.000000)
(99.000000, 0.000000)
 };
    \end{axis}
\end{tikzpicture}

%% file: figures/pseudoshape/alif_spikes.tex
\begin{tikzpicture}
    \begin{axis}[
      style={font=\scriptsize},
      width=9.3cm,
      height=3.5cm,
      xmin=0,
      xmax=100,
      ymax=1,
      ymin=0,
      minor y tick num=1,
      minor x tick num=1,
      legend pos=south east,
      xlabel={timesteps in \textbf{ms}},
      xticklabels={,,20,40,60,80,},
      yticklabels={,,0.2,,,0.8,},
      try min ticks=5,
      ylabel={\smash{$z_j^t$}},
    ]
        \addplot[mark=none,black] plot coordinates {
(0.000000, 0.000000)
(1.000000, 0.000000)
(2.000000, 0.000000)
(3.000000, 0.000000)
(4.000000, 0.000000)
(5.000000, 0.000000)
(6.000000, 0.000000)
(7.000000, 0.000000)
(8.000000, 0.000000)
(9.000000, 0.000000)
(10.000000, 0.000000)
(11.000000, 0.000000)
(12.000000, 0.000000)
(13.000000, 1.000000)
(14.000000, 0.000000)
(15.000000, 0.000000)
(16.000000, 0.000000)
(17.000000, 0.000000)
(18.000000, 0.000000)
(19.000000, 0.000000)
(20.000000, 0.000000)
(21.000000, 0.000000)
(22.000000, 0.000000)
(23.000000, 0.000000)
(24.000000, 0.000000)
(25.000000, 0.000000)
(26.000000, 0.000000)
(27.000000, 0.000000)
(28.000000, 0.000000)
(29.000000, 0.000000)
(30.000000, 0.000000)
(31.000000, 0.000000)
(32.000000, 0.000000)
(33.000000, 1.000000)
(34.000000, 0.000000)
(35.000000, 0.000000)
(36.000000, 0.000000)
(37.000000, 0.000000)
(38.000000, 0.000000)
(39.000000, 0.000000)
(40.000000, 0.000000)
(41.000000, 0.000000)
(42.000000, 0.000000)
(43.000000, 0.000000)
(44.000000, 0.000000)
(45.000000, 0.000000)
(46.000000, 0.000000)
(47.000000, 0.000000)
(48.000000, 0.000000)
(49.000000, 0.000000)
(50.000000, 0.000000)
(51.000000, 0.000000)
(52.000000, 0.000000)
(53.000000, 0.000000)
(54.000000, 0.000000)
(55.000000, 0.000000)
(56.000000, 0.000000)
(57.000000, 0.000000)
(58.000000, 1.000000)
(59.000000, 0.000000)
(60.000000, 0.000000)
(61.000000, 0.000000)
(62.000000, 0.000000)
(63.000000, 0.000000)
(64.000000, 0.000000)
(65.000000, 0.000000)
(66.000000, 0.000000)
(67.000000, 0.000000)
(68.000000, 0.000000)
(69.000000, 0.000000)
(70.000000, 0.000000)
(71.000000, 0.000000)
(72.000000, 0.000000)
(73.000000, 0.000000)
(74.000000, 0.000000)
(75.000000, 0.000000)
(76.000000, 0.000000)
(77.000000, 0.000000)
(78.000000, 0.000000)
(79.000000, 0.000000)
(80.000000, 0.000000)
(81.000000, 0.000000)
(82.000000, 0.000000)
(83.000000, 0.000000)
(84.000000, 0.000000)
(85.000000, 0.000000)
(86.000000, 0.000000)
(87.000000, 0.000000)
(88.000000, 0.000000)
(89.000000, 0.000000)
(90.000000, 0.000000)
(91.000000, 0.000000)
(92.000000, 0.000000)
(93.000000, 1.000000)
(94.000000, 0.000000)
(95.000000, 0.000000)
(96.000000, 0.000000)
(97.000000, 0.000000)
(98.000000, 0.000000)
(99.000000, 0.000000)
 };
    \end{axis}
\end{tikzpicture}